\documentclass[format=acmsmall,review=false]{acmart}

\makeatletter
\def\@ACM@checkaffil{
    \if@ACM@instpresent\else
    \ClassWarningNoLine{\@classname}{No institution present for an affiliation}%
    \fi
    \if@ACM@citypresent\else
    \ClassWarningNoLine{\@classname}{No city present for an affiliation}%
    \fi
    \if@ACM@countrypresent\else
        \ClassWarningNoLine{\@classname}{No country present for an affiliation}%
    \fi
}
\makeatother

\usepackage{acm-ec-25}
\usepackage{booktabs} 
\usepackage[ruled]{algorithm2e} 
\usepackage{listings}
\usepackage{xcolor}
\usepackage{subcaption}
\usepackage{tikz}
\usetikzlibrary{arrows.meta}
\usepackage{longtable}
\usepackage{makecell}
\usepackage{adjustbox}
\usepackage{colortbl}
\usepackage{tcolorbox}
\usepackage{graphicx}
\usepackage{placeins}
\tcbuselibrary{listingsutf8}

\SetAlFnt{\small}
\SetAlCapFnt{\small}
\SetAlCapNameFnt{\small}
\SetAlCapHSkip{0pt}
\IncMargin{-\parindent}

\setcitestyle{authoryear}

\title[The Effect of State Representation on LLM Agent Behavior in Dynamic Routing Games]{The Effect of State Representation on LLM Agent Behavior in Dynamic Routing Games}

\author{Lyle Goodyear}
\affiliation{%
  \institution{Stanford University}
  \city{Stanford}
  \state{CA}
  \country{USA}
}
\email{lyleg@stanford.edu}

\author{Rachel Guo}
\affiliation{%
  \institution{Stanford University}
  \city{Stanford}
  \state{CA}
  \country{USA}
}
\email{rsguo@stanford.edu}

\author{Ramesh Johari}
\affiliation{%
  \institution{Stanford University}
  \city{Stanford}
  \state{CA}
  \country{USA}
}
\email{rjohari@stanford.edu}

\begin{abstract}
Large Language Models (LLMs) have shown promise as decision-makers in dynamic settings, but their stateless nature necessitates creating a natural language representation of history. We present a unifying framework for systematically constructing natural language "state" representations for prompting LLM agents in repeated multi-agent games. Previous work on games with LLM agents has taken an ad hoc approach to encoding game history, which not only obscures the impact of state representation on agents' behavior, but also limits comparability between studies. Our framework addresses these gaps by characterizing methods of state representation along three axes: \emph{action informativeness} (i.e., the extent to which the state representation captures actions played); \emph{reward informativeness} (i.e., the extent to which the state representation describes rewards obtained); and \emph{prompting style} (or \emph{natural language compression}, i.e., the extent to which the full text history is summarized).

We apply this framework to a dynamic selfish routing game, chosen because it admits a simple equilibrium both in theory and in human subject experiments \cite{rapoport_choice_2009}. Despite the game's relative simplicity, we find that there are key dependencies of LLM agent behavior on the natural language state representation. In particular, we observe that representations which provide agents with (1) summarized, rather than complete, natural language representations of past history; (2) information about regrets, rather than raw payoffs; and (3) limited information about others' actions lead to 
behavior that more closely matches game theoretic equilibrium predictions, and with more stable game play by the agents.  By contrast, other representations can exhibit either large deviations from equilibrium, higher variation in dynamic game play over time, or both.

Taken together, these findings and our framework provide practical guidance for experimenters to systematically refine state representations when harnessing LLMs as strategic agents in dynamic settings.
\end{abstract}


\begin{document}

\begin{titlepage}


\maketitle

\setcounter{tocdepth}{1} 
\tableofcontents

\end{titlepage}

\section{Introduction}

Strategic decision-making is fundamental to understanding how agents interact in dynamic environments, from economic and social systems to robotics and artificial intelligence. In these settings, agents repeatedly make choices while adapting to others' behavior. Repeated games serve as a canonical framework for studying such interactions, capturing how agents best-respond to the previous actions of their peers. Traditionally, researchers have studied these processes analytically, with human subjects, or with reinforcement learning algorithms, provided carefully-selected state representations. However, recent advances in large language models (LLMs) have led to the development of a new class of decision-makers---the LLM-powered agent which calls upon an LLM for reasoning, acting according to the outputted decision \cite{yao_react_2023,horton_large_2023}. LLMs are inherently \textit{stateless}, instead relying on textual prompts to understand game history. This motivates the question: \textit{How should past interactions be encoded in natural language so that LLMs can effectively make decisions in strategic settings over time?}

Prior work on LLMs in dynamic games has taken an ad hoc approach to representing state, with authors using different representations across different games \cite{fontana_nicer_2024, akata_playing_2023, duan_gtbench_2024, park_llm_2024, xu_magic_2024, phelps_machine_2024}. Without a common framework for structuring state representation, it is difficult to systematically compare results across studies or identify aspects of state representation that have a critical impact on agent performance. We aim to address this gap in the literature by proposing a unifying framework for constructing \textit{natural language state representations}; see Section \ref{sec:Framework}. We characterize game-history prompts along three key axes: (1) \textit{action informativeness} (i.e., the granularity of information provided about agents' previous actions), (2) \textit{reward informativeness} (i.e., the granularity of information provided about agents' previous rewards), and (3) \textit{prompting style} (also referred to as \textit{natural language compression}, i.e., the extent to which relevant information about the previous chat history is summarized). By defining key dimensions along which natural language state representations can vary, our framework provides structure to the near-infinite space of natural language representations of state. This structure enables the experimental testing of the impact of variation along each axis of game state information on LLM decision-making.

In Section \ref{sec:game}, we apply this framework to a repeated selfish routing game on a network.  The network we consider is a classical selfish routing game that exhibits Braess's Paradox \cite{roughgarden_selfish_2005}. This game serves as an ideal testbed because it has a well-understood, simple equilibrium structure, and it enjoys extremely robust guarantees for learning behavior.  

Prior work on this game has shown that human subjects that play repeatedly approach equilibrium behavior \cite{rapoport_choice_2009} with enough rounds of play.  For our purposes, we adapt the experimental design of \cite{rapoport_choice_2009}---which implements an atomic version of the classical routing game with human participants---for use with LLM agents. In particular, we systematically vary the state representations provided to agents along the three axes of our framework: action informativeness, reward informativeness, and natural language compression. Some of these representations mirror those implicitly used in prior work on LLM game-playing, while others are novel variations designed to isolate the effects of each dimension on agent behavior.  We outline our architectural approach in Section \ref{sec:architecture}.  We study the behavior of our agents \textit{experimentally} by simulating game play exactly as in the lab experiment of \cite{rapoport_choice_2009}.  A quantitative description of our evaluation approach is provided in Section \ref{sec:metrics}.

We report on our findings in Section \ref{sec:results}.  Our empirical findings suggest that the design of state representations has a significant impact on LLM agents' learning dynamics and convergence to equilibrium.  We focus in particular on how state representations lead agent behavior to be closer or further from theoretical equilibrium behavior.  Our main findings are as follows:
\begin{enumerate}
    \item {\em Summary vs.~full chat prompting style in the state representation.}  Our results strongly indicate that agents' behavior more closely matches equilibrium behavior when they are provided with appropriate natural language {\em summarizations} of the full chat history of previous game play, rather than the full chat history itself.  This finding likely stems from the simplification and structure that results in the context provided to the LLM agents. 
    \item {\em Regret vs.~payoff in the state representation.} Our results indicate that providing agents with their regret (i.e., gap in payoff to the optimal action) rather than their raw payoff in each round leads to more stable behavior, which also more closely matches equilibrium.  This finding suggests that agents benefit from information that allows them to discern how to change their action, rather than just the outcome of the action they chose.
    \item {\em Own actions vs.~everyone's actions in the state representation.}  Finally, our findings suggest that providing {\em less} information about others' actions leads to more stable agent behavior, i.e., fewer individuals changing their action choices over time, and thus closer matches to equilibrium behavior.
\end{enumerate}

In Section \ref{sec:discussion}, we provide some discussion of potential qualitative explanations for our quantitative findings, as well as directions for future research.




\section{Related Works}
\label{sec:related}

\subsection{Simulating Human Behavior with LLMs}
LLMs exhibit emergent capabilities that capture nuances of human behavior, a result of the massive corpora of text used to train these models. This has sparked research into the potential of LLMs for simulating human social interactions, biases, and economic decision-making. \cite{horton_large_2023} argues that LLMs can be treated as an implicit computational model of humans---a "homo silicus"---which can be endowed with preferences and used to study human economic behavior. Similarly, significant work has been done exploring whether LLMs can simulate well-understood human economic biases, such as temporal discounting or risk aversion \cite{coletta_llm-driven_2024, leng_can_2024, kitadai_can_2024, ross_llm_2024}. Furthermore, some authors have focused on endowing LLMs with various demographic traits, exploring the capacity of LLMs as replacements for human in social science research \cite{aher_using_2023, argyle_out_2023}. Finally, \cite{park_generative_2023} proposed highly granular social simulations powered by LLM agents. Our work extends research into LLM-powered social and economic simulations by implementing a multi-agent selfish routing game with LLMs, a previously unexplored domain.

\subsection{LLMs Playing Games}
A growing body of research evaluates LLMs as rational agents in game-theoretic scenarios. While much of this work has focused on static two-player games, such as the \textit{prisoner's dilemma} \cite{herr_are_2024, guo_gpt_2023} and other classical two-player games \cite{shapira_glee_2024, wang_tmgbench_2024, jia_large_2025}, fewer studies have examined games with more than two players. Some research has explored complex text-based games like \textit{Werewolf} \cite{xu_exploring_2024}, \textit{Avalon} \cite{lan_llm-based_2024}, and \textit{Guandan} \cite{yim_evaluating_2024}, where multiple agents interact via natural language. However, these games lack well-defined equilibria, making it difficult to systematically evaluate LLM strategic decision-making. More relevant to our work, \cite{huang_how_2024} examines 10 structured multi-agent games, including two sequential settings. 

Regarding LLMs and \textit{repeated} games, most of the literature has focused on the iterated prisoner's dilemma \cite{duan_gtbench_2024,fontana_nicer_2024,park_llm_2024,phelps_machine_2024,xu_magic_2024} with conflicting results; for example, \cite{akata_playing_2023} found that GPT-4 was particularly unforgiving during play, while \cite{fontana_nicer_2024} found that Llama2 was more willing to cooperate than human subjects. A few studies have extended this work to other games like \textit{battle of the sexes} \cite{akata_playing_2023}, \textit{public goods} games \cite{xu_magic_2024}, and general matrix games \cite{park_llm_2024}. \cite{yu_llm-based_2025} show that augmenting LLM agents with explicit models of opponent behavior improves performance in dynamic multi-agent games, highlighting the influence of structured history representation on agent behavior.

Across these studies, the effect of state representation on agent behavior has not been systematically studied; each work has its own desiderata that impact the choice of state representation. In particular, these works take varied, ad hoc approaches to encoding state information in the prompt, ranging from tabular summaries of previous interactions \cite{fontana_nicer_2024, akata_playing_2023, duan_gtbench_2024, park_llm_2024} to chatbot-style interaction between the game environment and agents \cite{xu_magic_2024, phelps_machine_2024}. There is no standardization in these methods and little understanding of how the encoding of historical information impacts LLM decision-making, making comparisons across results difficult. In Section \ref{sec:Framework}, we provide a framework to reason about the space of possible state representations, providing key dimensions on which to experiment.

\subsection{LLMs, Learning, and Games}
LLMs are built on the transformer architecture \cite{vaswani_attention_2023} which enables them to exhibit \textit{in-context learning} \cite{brown_language_2020}, the ability to adapt to new tasks based on contextual information in the prompt without updating model parameters. This emergent ability has been shown to approximate a number of machine learning algorithms entirely in-context \cite{bai_transformers_2023}, suggesting that LLMs may possess implicit learning mechanisms relevant to dynamic decision-making. Prior work has investigated the \textit{regret} of LLM agents in online learning and repeated games \cite{park_llm_2024}. We build on this work, examining how the learning dynamics and regret behavior of LLM agents are influenced by the choice of history encoding, i.e., the state representation.

\section{State Representations for LLM Game-Playing Agents: A Framework}
\label{sec:Framework}

In this paper, we consider a number of identical LLM agents playing a dynamic repeated game.  For an agent to play the game, in each round it needs a natural language context that provides historical information about the play of the game to inform the action choice in the current round.  From the vantage point of an agent, we define a {\em state representation} as this structured natural language encoding of past interactions.

In principle, the natural language representation of the history of a game can rapidly become intractably large, even for simple games and few rounds. Further, as illustrated by prior work (see Section \ref{sec:related}), there are many choices possible for such representations, and little structured guidance on how to choose between them.  With these observations as motivation, in this section we introduce a structured framework to organize our investigation of state representations.  This framework then informs our subsequent study of LLM game-playing agents in the remainder of the paper.

In our framework, we choose to focus on variation in state representations along three particularly salient primary ``axes'': (1) {\em action informativeness}; (2) {\em reward informativeness}; and (3) natural language compression, or what we refer to succinctly as {\em prompting style}. Each axis captures a different dimension of the historical information supplied to the agent.  We now describe each in turn below.

\paragraph{(1) Action Informativeness.}  This aspect of a state representation specifies {\em which agents}’ actions are included in the historical record. In our experiments, we contrast representations that include only (i) {\em the focal agent’s \em own past actions} with those that \textit{also} incorporate the (ii) {\em actions of other agents} (We note that in the literature on algorithms for no-regret learning in games, algorithms typically depend on the actions of others only through the resulting payoffs; see, e.g., \cite{cesa-bianchi_prediction_2006}).  

There are at least two reasons that this is an important axis of variation. First is inference regarding peer behavior: An agent that observes only its own actions may have a limited view of the strategic environment and could struggle to infer the behavior or intentions of its peers. In contrast, including the actions of all agents can provide richer context, enabling the agent to better adjust its strategy in response to collective dynamics.
Second is the implication of the granularity of feedback: The decision of whether to include external actions may influence the internal reasoning of the agent. In settings where strategic adaptation is critical, richer action information could lead to more effective learning; however, it also increases the complexity of the context the agent must process.  Given the multiple potential impacts of action informativeness on an LLM agent's strategic reasoning, the choice along this axis may have significant implications for both the speed and stability of convergence to equilibrium.

\paragraph{(2) Reward Informativeness.}  The reward informativeness axis determines the type of performance feedback provided to the agent. In our framework, agents receive either historical (i) {\em payoffs} or (ii) {\em regret}.  In the first case, the agent is given the history of accumulated rewards obtained from past actions.  In the second case, the agent is provided with counterfactual performance measures that indicate how much better the agent could have done by choosing optimal actions.  

The distinction between payoffs and regrets is informed by the online learning literature. Many algorithms, e.g., those based on multiplicative weights (e.g., \cite{freund_adaptive_1999, arora2012multiplicative} or follow-the-regularized-leader (FTRL, \cite{shalev2007online,abernethy2008competing}), rely on payoff-based feedback to adjust strategies.  By contrast, regret-based methods, e.g., regret matching (\cite{hart2013simple, cesa-bianchi_prediction_2006}) and related methods, use regrets to guide decision-making.  Although these different types of feedback have been studied in algorithmic learning, their impact on the internal strategic reasoning of LLM agents remains less clear. Providing regret information might enhance an agent’s ability to differentiate between optimal and suboptimal choices, potentially accelerating convergence, but it also introduces a counterfactual reasoning challenge that could alter how the LLM internally represents and responds to its past performance.  Again, these potential tradeoffs can impact the speed and stability of convergence to equilibrium.

We note that a related distinction arises in the literature on algorithmic online learning in games, namely, the distinction between {\em full} and {\em bandit} feedback (see, e.g., \cite{cesa-bianchi_prediction_2006, lattimore2020bandit}).  In algorithms that act on full feedback (such as Multiplicative Weights \cite{freund_adaptive_1999}), the algorithm is provided with the full vector of payoffs as a function of the agent's action at each time period.  By contrast, in algorithms that act on bandit feedback (also sometimes called ``partial feedback''), the algorithm only observes the realized payoff for actions that are played; this is the case in algorithms such as EXP3 \cite{auer_nonstochastic_2002}.  In our formulation, payoff-based feedback is analogous to bandit-feedback.  Regret-based feedback is less informative than full feedback, but more reasonable when natural language context is provided to the agent (as, in particular, it eliminates the computational step of requiring the agent to reason about the optimal action at each stage).  


\paragraph{(3) Prompting Style.}  The third axis concerns natural language compression, i.e., how historical information is structured and presented within the prompt.   Clearly, this is an essential aspect of state representation for LLM agents.  It is also potentially very broad, so we structure our investigation by focusing on the distinction between two main prompting styles: (i) {\em full-chat prompting} and (ii) {\em summarized prompting}.

In full-chat prompting, the LLM agent is provided with a sequential transcript of the game, including a system message (which explains the game and expected response format), the agent’s previous responses, and round-by-round results. At the end of each round, a new message summarizing the most recent outcomes is appended to the dialogue history. This approach mimics a real-time chat interface, offering a detailed, uncompressed view of the game’s progression.

In summarized prompting, the LLM agent receives a compressed summary of historical interactions. Instead of a complete transcript, the prompt includes the system message and a cumulative summary that aggregates the outcomes from all previous rounds—typically presented in a tabular format. This method reduces context window usage and enforces a more structured, meta-learning approach by filtering out less critical details.

Prior work on LLMs in dynamic games has predominantly employed one of these two prompting styles (e.g., full-chat \cite{xu_magic_2024, phelps_machine_2024} or summarized \cite{fontana_nicer_2024, akata_playing_2023, duan_gtbench_2024, park_llm_2024}). Moreover, recent findings suggest that as prompts approach the context window limit, LLMs struggle to extract relevant information \cite{liu_lost_2024}. Hence, we are led to investigate whether reducing context length via summarization may facilitate more effective inference and decision-making by the agent.

\paragraph{Remark: Historical Memory.}  We remark that in our state representation, we retain the entire history, rather than truncating the history at a given depth (e.g., only a fixed number of previous rounds of information).  There are two reasons for this choice; since it is not feasible to investigate all dimensions of state representation variation at once, we chose to focus on those we felt are most relevant to distinguishing performance of LLM agent behavior in dynamic repeated games.  Second, in our particular experiments the entire history can be included within the available context window.  Investigating the role of history truncation is an interesting direction for future research.

\paragraph{Notation.} 
To systematically assess the three axes of state representations that we have delineated, in our experiments we test all possible combinations, yielding eight unique state representations. Each representation is denoted using a naming convention that encodes:
\begin{itemize}
    \item {\em Action informativeness}: \textbf{O} if only an agent's own action history is included; \textbf{E} if all agents' action history is included;
    \item {\em Reward informativeness}: \textbf{P} for payoff-based feedback; \textbf{R} for regret-based feedback; and
    \item {\em Prompting style}: \textbf{F} for full-chat, \textbf{S} for summarized.
\end{itemize}

\begin{figure}[ht]
    \centering
    \begin{subfigure}[t]{0.49\linewidth}
        \centering
        \begin{tcolorbox}[colframe=red!60, colback=red!10, arc=10pt, left=2pt, right=2pt, boxrule=1pt, width=\dimexpr\linewidth-4pt\relax, title={Game Environment (Round 1)}]\footnotesize
            You will be participating in an experiment on route selection in traffic networks $\dots$ The available routes are: O-L-D, O-R-D
        \end{tcolorbox}

        \begin{tcolorbox}[colframe=gray!60, colback=gray!10, arc=10pt, left=2pt, right=2pt, boxrule=1pt, width=\dimexpr\linewidth-4pt\relax, title={LLM}]\footnotesize
            To make an informed decision, let's analyze the potential costs $\dots$ Based on this assessment, I will choose: "O-R-D"
        \end{tcolorbox}

        \begin{tcolorbox}[colframe=red!60, colback=red!10, arc=10pt, left=2pt, right=2pt, boxrule=1pt, width=\dimexpr\linewidth-4pt\relax, title={Game Environment (Round 2)}]\footnotesize
            Summary of previous round:
            
            Your Choice: O-R-D
            
            Route Choice Distribution: \{O-R-D: 13, O-L-D: 5\}

            Your Payoff: 60
        \end{tcolorbox}

        \begin{center}
            $\vdots$
        \end{center}

        \begin{tcolorbox}[colframe=gray!60, colback=gray!10, arc=10pt, left=2pt, right=2pt, boxrule=1pt, width=\dimexpr\linewidth-4pt\relax, title={LLM}]\footnotesize
            Let's analyze $\dots$ Given this analysis and aiming to optimize my payoff based on the previous round's distribution: "O-R-D"
        \end{tcolorbox}

        \begin{tcolorbox}[colframe=red!60, colback=red!10, arc=10pt, left=2pt, right=2pt, boxrule=1pt, width=\dimexpr\linewidth-4pt\relax, title={Game Environment (Round 10)}]\footnotesize
            Summary of previous round:
            
            Your Choice: O-R-D
            
            Route Choice Distribution: \{O-R-D: 17, O-L-D: 1\}
            
            Your Payoff: 20
        \end{tcolorbox}

        \caption{Fig \ref{fig:full_chat_prompt_example}. Example \textit{full-chat} representation given to an agent at the start of round 10. LLM is prompted with a list of the previous messages in the exchange.}
        \label{fig:full_chat_prompt_example}
    \end{subfigure}
    \hfill
    \begin{subfigure}[t]{0.49\linewidth}
        \centering
        \begin{tcolorbox}[colframe=blue!60, colback=blue!10, arc=10pt, left=2pt, right=2pt, boxrule=1pt, width=\dimexpr\linewidth-4pt\relax, title={Game Environment (Round 10)}]\footnotesize
            You will be participating in an experiment on route selection in traffic networks $\dots$ The available routes are: O-L-D, O-R-D
        \end{tcolorbox}

        \begin{tcolorbox}[colframe=blue!60, colback=blue!10, arc=10pt, left=2pt, right=2pt, boxrule=1pt, width=\dimexpr\linewidth-4pt\relax, title={Game Environment (Round 10)}]\footnotesize
            Summary of previous rounds:
            
              Round 1:
              
                Your Choice: O-R-D
                
                Route Choice Distribution: \{O-R-D: 13, O-L-D: 5\}
                
                Your Payoff: 60
                
            \begin{center}
                $\vdots$
            \end{center}
              Round 7:
              
                Your Choice: O-R-D
                
                Route Choice Distribution: \{'O-R-D': 12, 'O-L-D': 6\}
                
                Your Payoff: 70
                
              \bigskip  
              Round 8:
              
                Your Choice: O-L-D
                
                Route Choice Distribution: \{'O-L-D': 11, 'O-R-D': 7\}
                
                Your Payoff: 80
                
              \bigskip    
              Round 9:
              
                Your Choice: O-R-D 
                
                Route Choice Distribution: \{O-L-D: 17, O-R-D: 1\}

                Your Payoff: 20
        \end{tcolorbox}

        \caption{Fig \ref{fig:summary_prompt_example}. Example \textit{summary} representation given to an agent at the start of round 10. LLM is prompted with a summary of its previous interactions.}
        \label{fig:summary_prompt_example}
    \end{subfigure}
    \caption{Comparison of prompt given to agents in \textit{full-chat} (Fig. \ref{fig:full_chat_prompt_example}) and \textit{summarized} (Fig. \ref{fig:summary_prompt_example}) representations.}
    \label{fig:chat_comparison}
    \Description{}
\end{figure}

\begin{figure}[ht]
    \centering
    \begin{subfigure}{0.42\textwidth} 
        \centering
        \setlength\tabcolsep{2pt} 
        \renewcommand{\arraystretch}{0.8} 
        \footnotesize 
        \begin{tabular}{@{}p{1cm} >{\raggedright\arraybackslash}p{1.4cm} >{\raggedright\arraybackslash}p{1.7cm} >{\raggedright\arraybackslash}p{1.3cm}@{}}
            \toprule
            \textbf{Rep.} & \textbf{Prompting} & \textbf{Action Info} & \textbf{Reward Info} \\
            \midrule
            \rowcolor[HTML]{F3F3F3} F-PO & Full-Chat & Own + Others' & Payoff \\
            S-PO & Summary & Own + Others' & Payoff \\
            \rowcolor[HTML]{F3F3F3} F-RO & Full-Chat & Own + Others' & Regret \\
            S-RO & Summary & Own + Others' & Regret \\
            \rowcolor[HTML]{F3F3F3} F-R  & Full-Chat & Own Only      & Regret \\
            S-R  & Summary   & Own Only      & Regret \\
            \rowcolor[HTML]{F3F3F3} F-P  & Full-Chat & Own Only      & Payoff \\
            S-P  & Summary   & Own Only      & Payoff \\
            \bottomrule
        \end{tabular}
        \caption{Fig.~\ref{fig:framework_table}. Summary of state representations tested.}
        \label{fig:framework_table}
    \end{subfigure}
    \hfill
    \begin{subfigure}{0.54\textwidth} 
        \centering
        \includegraphics[width=\textwidth]{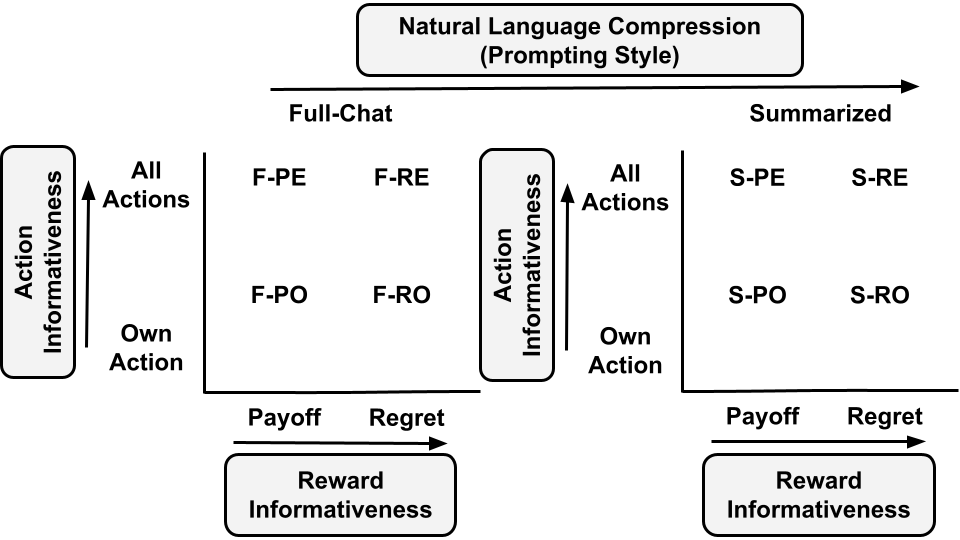} 
        \caption{Fig.~\ref{fig:framework_diagram}. Visual comparison of axes of informativeness.}
        \label{fig:framework_diagram}
    \end{subfigure}
    \label{fig:table_diagram}
    \Description{}
\end{figure}

\section{Methods: Game, Experiment Design, and Evaluation}
\label{sec:methods}

\subsection{Dynamic Routing Game}
\label{sec:game}

Our investigation of state representations is grounded in the study of a repeated {\em atomic selfish routing} game \cite{roughgarden_selfish_2005, roughgarden2004selfish_atomic, kontogiannis2005atomic} (We consider atomic games since we have finitely many players; in nonatomic games, agents are infinitesimal).  In such games, multiple agents seek to minimize their individual travel costs from source to destination by selecting routes in a network with congestion-dependent edge costs.  Formally, in a static atomic selfish routing game, a total of $n$ agents select routes simultaneously and experience cost equal to the sum of the congestion on each edge in their selected route.  In our setting, we consider dynamic repetition of such an atomic game over $T$ rounds.

We consider a specific (canonical \cite{roughgarden_selfish_2005}) instantiation of the selfish routing game, described by the two networks in Figure \ref{fig:networks} For simplicity, suppose $n$ is even. In each figure, the agents' goal is to traverse from $O$ to $D$.
In Figure \ref{fig:network-A}, agents can choose either the ``upper'' route (\textit{O-L-D}) or the ``lower'' route (\textit{O-R-D}).  Let $n_L$ (resp., $n_R$) denote the number of agents choosing the upper (resp., lower) route in a particular action configuration, with $n_L + n_R = n$.  Each edge in the figure is labeled with a {\em congestion cost function}, as a function of the total number of agents on the edge.  Thus, any agent choosing the upper (resp., lower) route incurs a cost $10 n_L + 210$ (resp., $10 n_R + 210$). Figure \ref{fig:network-B} is identical, except for the introduction of a third ``bridge'' route: \textit{O-L-R-D}.  If $n_L, n_B, n_R$ are the number of agents choosing the upper, bridge, and lower routes respectively, then an agent traversing the bridge route incurs a total cost $10(n_L + n_B) + 10(n_R + n_B)$.  We refer to the left game as {\em Game A}, and the right game as {\em Game B}.  

It is well known that the comparison between the static equilibria of the two games with the structure in Figure \ref{fig:networks} exhibits {\em Braess' paradox} (BP) in general, both in nonatomic and atomic variants: i.e., there exist configurations where the equilibrium cost to agents in the right game is {\em higher} than the equilibrium cost in the left game, despite the presence of an additional edge and route \cite{roughgarden_selfish_2005, correa2011wardrop}.  This is a version of the game theoretic insight in the simple prisoner's dilemma, where equilibrium behavior need not be Pareto efficient.

In our paper, we investigate a dynamic, repeated version of these two games.  
Our experimental setup is chosen to match the study of \cite{rapoport_choice_2009}, who implement both Game A and Game B in a lab experiment with $18$ human subjects.\footnote{Relative to their paper, we changed the node labels from $\{O,A,B,D\}$ to $\{O,L,R,D\}$ to reduce bias present in LLM route choices \cite{zheng_large_2024}.}  It is straightforward to check that if $n = 18$, then in the left game the pure strategy Nash equilibria consist of any configuration where  $n_L = n_R = 9$, and all these equilibria yield identical costs to each player of $300$.  In the right game the unique pure strategy Nash equilibrium is $n_L = n_R = 0$, and $n_B = 18$ (in fact, the bridge route is a weak dominant strategy); each player incurs a cost of $360$.   The fact that this is higher than $300$ is evidence of Braess' paradox.  

Because both games possess unique pure strategy Nash equilibrium payoffs, it is straightforward to check that both repeated games also have unique pure strategy subgame perfect Nash equilibrium payoffs.  In particular, in any SPNE of the finitely repeated Game A, players must split evenly at every stage and earn payoffs of $300$ at every stage.  In the unique SPNE of the finitely repeated Game B, all players must choose route \textit{O-L-R-D} at every stage (We note in passing that the left game also possesses a unique mixed strategy Nash equilibrium where agents uniformly randomize between the two routes). Note also that this dynamic repeated selfish routing game is a potential game \cite{monderer1996potential}, so it is well known that {\em best response dynamics} converge to a pure strategy Nash equilibrium.  Best response dynamics involve each agent myopically choosing a best response from the last round of play.

Taken together, the strong characterizations of static and dynamic equilibria, as well as robust dynamic learning-theoretic of these two games, make them ideal targets for investigation of state representations.  Because we understand equilibrium and off-equilibrium behavior well for these games, we have strong benchmarks for comparison.

\cite{rapoport_choice_2009} repeats each game for $T = 40$ rounds with $n = 18$ human subjects. They demonstrate that human subjects converge to the game's theoretical Nash equilibria in laboratory experiments.  Given this finding, and given that our goal is to study the convergence properties of LLM agents, we use a similar experiment design, cost functions, and number of agents ($n=18$) as in \cite{rapoport_choice_2009}. This approach also has the benefit of enabling comparisons between LLM agent behavior, human subject behavior, and theoretical equilibrium agent behavior in the game.  Note that in \cite{rapoport_braess_2008}, experiments were run such that Game B immediately followed Game A---or the reverse---with the same participants; we instead split Game A and Game B into separate self-contained experiments to reduce the cost of API calls. \cite{rapoport_choice_2009} found no difference in behavior depending on which network was presented first.

\begin{figure}[ht]
    \centering
    \begin{subfigure}{0.45\textwidth}
        \centering
        \begin{tikzpicture}[scale=0.90, node distance=2cm]
            \node[draw, circle] (O) at (0,0) {O};
            \node[draw, circle] (L) at (3,2) {L};
            \node[draw, circle] (R) at (3,-2) {R};
            \node[draw, circle] (D) at (6,0) {D};

            \draw[->, thick] (O) -- (L) node[midway, above, yshift=5pt] {$10x$};
            \draw[->, thick] (O) -- (R) node[midway, below, yshift=-5pt] {$210$};
            \draw[->, thick] (L) -- (D) node[midway, above, yshift=5pt] {$210$};
            \draw[->, thick] (R) -- (D) node[midway, below, yshift=-5pt] {$10x$};
        \end{tikzpicture}
        \caption{Fig. \ref{fig:network-A}. Network in Game $A$}
        \label{fig:network-A}
    \end{subfigure}
    \hfill
    \begin{subfigure}{0.45\textwidth}
        \centering
        \begin{tikzpicture}[scale=0.90, node distance=2cm]
            \node[draw, circle] (O) at (0,0) {O};
            \node[draw, circle] (L) at (3,2) {L};
            \node[draw, circle] (R) at (3,-2) {R};
            \node[draw, circle] (D) at (6,0) {D};

            \draw[->, thick] (O) -- (L) node[midway, above, yshift=5pt] {$10x$};
            \draw[->, thick] (O) -- (R) node[midway, below, yshift=-5pt] {$210$};
            \draw[->, thick] (L) -- (D) node[midway, above, yshift=5pt] {$210$};
            \draw[->, thick] (R) -- (D) node[midway, below, yshift=-5pt] {$10x$};
            \draw[->, thick] (L) -- (R) node[midway, right, xshift=10pt] {0};
        \end{tikzpicture}
        \caption{Fig. \ref{fig:network-B}. Network in Game $B$}
        \label{fig:network-B}
    \end{subfigure}
    \caption{Comparison of networks used in Game $A$ and Game $B$, where $x$ denotes the number of agents on a given edge.}
    \label{fig:networks}
    \Description{}
\end{figure}

\subsection{Game Environment and LLM Agent Architecture}
\label{sec:architecture}

In our simulation, $n = 18$ LLM-based agents play Games A and B repeatedly for $T = 40$ rounds.  This section describes the architecture of the LLM agents, the game environment, and the prompting methods used in our simulation. Our implementation is built using LangChain\footnote{\href{https://www.langchain.com/}{https://www.langchain.com/}}, OpenAI's \texttt{gpt-4o-2024-08-06}, and a custom traffic network simulation framework.

\paragraph{Game Environment.} The congestion network is represented as a directed graph with associated cost functions. In each round, the game environment iteratively prompts decision from each agent. It then stores each agent's decision, payoff (computed by evaluating cost functions in the network), and regret (computed for a given agent by fixing their peers' decisions and finding counterfactual payoffs for each other route). This information is used to prompt agents in the next round. The network is reset at the end of each round.

\paragraph{LLM Agents.} Each agent is instantiated as a \texttt{gpt-4o-2024-08-06} session in LangChain. We set the temperature of GPT-4o to 1 throughout experiments to induce varied behavior across runs. Regardless of the state representation, every time an LLM agent is prompted, it receives a description of the game which is directly adapted from the text instructions given to subjects in \cite{rapoport_choice_2009}. In each round, the prompts are dynamically generated based on the focal agent and state representation. Agents are given historical information formatted to the chosen state representation (the same across all agents in a single game). Agents are then iteratively prompted with the above information and a request to provide a route selection from the available options (differing in Game A and Game B) in a JSON format. Once decisions have been collected and processed, the next round begins.

\paragraph{Prompting.} Each prompt for an agent decision is a separate API call with the information relevant to that agent only. Beyond the included information, the primary difference between prompt structures across representations is \textit{prompting style}, detailed in Section \ref{sec:Framework} (in particular, Figures \ref{fig:full_chat_prompt_example} and \ref{fig:summary_prompt_example}). Each prompt for an agent decision ends with a request to think step-by-step, intended to elicit chain-of-thought reasoning (introduced by \cite{wei_chain--thought_2022}) where agents visibly decompose their thinking into small steps. This increases the depth of reasoning while also allowing us to observe an agent's thought process in natural language. We provide examples of the full historical information text in every summarized representation in Figures \ref{fig:spe-example} (S-PE), \ref{fig:spo-example} (S-PO), \ref{fig:sre-example} (S-RE), and \ref{fig:sro-example} (S-RO). In the full-chat variants, the same information also includes LLM responses; Figures \ref{fig:full_chat_prompt_example} and \ref{fig:summary_prompt_example} display the differences in prompting style for F-PE and F-PO.

\begin{figure}[ht]
    \centering
    \begin{subfigure}[t]{0.49\linewidth}
        \centering
        \begin{tcolorbox}[colframe=red!60, colback=red!10, arc=10pt, left=2pt, right=2pt, boxrule=1pt, width=\dimexpr\linewidth-4pt\relax, title={Game Environment (Round 4)}]\footnotesize
            You are agent 0.
            
            \medskip
            
Summary of previous rounds:

\bigskip

  Round 1:

 Your Choice: O-R-D

    Route Choice Distribution: \{'O-R-D': 13, 'O-L-D': 5\}

    Your Payoff: 60

\bigskip

  Round 2:

    Your Choice: O-L-D
    
    Route Choice Distribution: \{'O-L-D': 17, 'O-R-D': 1\}
    
    Your Payoff: 20

\bigskip
  
  Round 3:
  
    Your Choice: O-R-D
    
    Route Choice Distribution: \{'O-R-D': 14, 'O-L-D': 4\}
    
    Your Payoff: 50
    
        \end{tcolorbox}
        \caption{Fig. \ref{fig:spe-example}. S-PE example round message.}
        \label{fig:spe-example}
    \end{subfigure}
    \hfill
    \begin{subfigure}[t]{0.49\linewidth}
        \centering
        \begin{tcolorbox}[colframe=red!60, colback=red!10, arc=10pt, left=2pt, right=2pt, boxrule=1pt, width=\dimexpr\linewidth-4pt\relax, title={Game Environment (Round 4)}]\footnotesize
You are agent 0.

\medskip

Summary of previous rounds:

\bigskip

  Round 1:
  
    Your Choice: O-R-D
    
    Route Choice Distribution: \{'O-R-D': 11, 'O-L-D': 7\}
    
    Your Regret: 30

\bigskip
    
  Round 2:
  
    Your Choice: O-L-D
    
    Route Choice Distribution: \{'O-L-D': 17, 'O-R-D': 1\}
    
    Your Regret: 150

\bigskip

  Round 3:
  
    Your Choice: O-R-D
    
    Route Choice Distribution: \{'O-R-D': 16, 'O-L-D': 2\}
    
    Your Regret: 130
    
    \end{tcolorbox}

        \caption{Fig. \ref{fig:sre-example}. S-RE example round message.}
        \label{fig:sre-example}
    \end{subfigure}

    \label{fig:examples1}
    \Description{}
\end{figure}

\begin{figure}[ht]
    \centering
    \begin{subfigure}[t]{0.49\linewidth}
        \centering
        \begin{tcolorbox}[colframe=red!60, colback=red!10, arc=10pt, left=2pt, right=2pt, boxrule=1pt, width=\dimexpr\linewidth-4pt\relax, title={Game Environment (Round 4)}]\footnotesize
You are agent 0.

\medskip

Summary of previous rounds:

\bigskip

  Round 1:
  
    Your Choice: O-R-D
    
    Your Payoff: 30

\bigskip   
    
  Round 2:
  
    Your Choice: O-L-D
    
    Your Payoff: 20

\bigskip
    
  Round 3:
  
    Your Choice: O-L-D
    
    Your Payoff: 180
    
    \end{tcolorbox}
        \caption{Fig. \ref{fig:spo-example}. S-PO example round message.}
        \label{fig:spo-example}
    \end{subfigure}
    \hfill
    \begin{subfigure}[t]{0.49\linewidth}
        \centering
        \begin{tcolorbox}[colframe=red!60, colback=red!10, arc=10pt, left=2pt, right=2pt, boxrule=1pt, width=\dimexpr\linewidth-4pt\relax, title={Game Environment (Round 4)}]\footnotesize
            You are agent 0.

            \medskip
            
            Summary of previous rounds:
            
            \bigskip
            
              Round 1:
              
                Your Choice: O-R-D
                
                Your Regret: 70
                
            \bigskip
            
              Round 2:
              
                Your Choice: O-L-D
                
                Your Regret: 150

            \bigskip
                
              Round 3:
              
                Your Choice: O-R-D
                
                Your Regret: 150   
                
        \end{tcolorbox}

        \caption{Fig. \ref{fig:sro-example}. S-RO example round message.}
        \label{fig:sro-example}
    \end{subfigure}

    \label{fig:examples2}
    \Description{}
\end{figure}

\subsection{Evaluation Metrics}
\label{sec:metrics}

Here we introduce the key metrics and visualizations that will be used to evaluate agent behavior. These metrics are computed separately for Game A and Game B, and provide insight into equilibrium convergence, stability, and learning dynamics. We also describe the visualizations we will use to track agent behavior over time.

For each representation and each trial (i.e., run of a repeated game among the corresponding LLM agents), we compute four {\em aggregate} metrics by averaging over all the rounds of the trial.  These metrics help us assess the extent to which agents' play conforms to static equilibrium behavior.  The metrics are defined as follows:
\begin{enumerate}
    \item {\em Mean Number of Agents on a Focal Route.} In particular, for Game B, we compute the mean number of agents on route \textit{O-L-R-D} in each round; this is the weak dominant strategy route.  For Game A, the calculation is adjusted to account for network symmetry. We first compute the of the number of agents on the \textit{least-congested} of the two routes ($\min(n_\textit{O-L-D}, n_\textit{O-R-D})$) in each round. We then average over the trial.  
    \item {\em Mean Payoff per Agent.}  We compute the mean of the payoffs of the agents in each round, then average over the trial.  
    \item {\em Mean Regret per Agent.}  Regret is the difference between the the best possible payoff in hindsight, and the received payoff, for a given agent in a given round.  We compute average regret across all agents in each round, then average over the trial.
    \item {\em Mean Number of Switches per Agent.}  We define a \textit{switch} for an agent if she chooses a different route in round $t$ than in round $t-1$.  In each trial, for each agent, we count their number of switches over the course of the game (where the max value for an agent is 39). We then average each agent's switch count within the trial.
\end{enumerate}
Each of these metrics are then averaged over trials, with a corresponding standard error computed over trials.

While aggregate statistics provide a summary of agent behavior, they do not capture how strategies evolve over time.  For this purpose, we also visuailze the metrics above on a per-round basis; for each representation and each round, we average over trials (and compute a corresponding standard error).\footnote{For the mean number of agents on a focal route, we make the same distinction as above; in Game A, we compute the average number of agents on the least-congested route, and in Game B, we compute the average number of agents on \textit{O-L-R-D}.}

Finally, to quantify how agent strategies evolve over time, we compute the {\em Kendall rank correlation coefficient} ($\tau$) between the round number and the \textit{equilibrium deviation score}, defined as the total distance between the number of agents on each route and the equilibrium prediction. Formally, Game A's deviation score $d_A$ in a given round is given by:
\begin{equation}
d_A=|n_{\textit{O-L-D}}-9|+|n_{\textit{O-R-D}}-9|
\end{equation}
(Recall that in Game A, 9 agents choose each route in Nash equilibrium.)
Similarly, Game B's deviation score $d_B$ in a given round is given by:
\begin{equation}
d_B=|n_{\textit{O-L-D}}-0|+|n_{\textit{O-R-D}}-0|+|n_{\textit{O-L-R-D}}-18|
\end{equation}
(Recall that in Game B, 18 agents choose route \textit{O-L-R-D} in Nash equilibrium.)  With these definitions $\tau$ for Game A (resp., Game B) is the correlation between the vector of $d_A$ (resp., $d_B$) in each round, and the vector $(1, 2, \ldots, 40)$ (i.e., the vector of round numbers).
Intuitively, a negative correlation $\tau$ indicates that the distance between agent choices and equilibrium predictions is decreasing over time, indicating convergence to equilibrium. Similarly, no correlation indicates no convergence, while positive correlation indicates that agents are performing further from equilibrium over time. The $\tau$ metric was also employed by \cite{rapoport_choice_2009} to examine the differences between equilibrium convergence between Game A and Game B. Note that a more negative $\tau$ does not necessarily mean \textit{faster} convergence; rather, it indicates a consistent decrease in the deviation score. This is an important nuance: $\tau$ specifically asks---conditional on being far from equilibrium in early rounds---have agents arrived monotonically arrived at equilibrium by the end of the game? The metric is \textit{not} informative if either agents \textit{start} at equilibrium or they \textit{never} reach equilibrium.

\section{Results}
\label{sec:results}

\subsection{Aggregate Behavior: Key Findings}
\label{sec:agg-metrics}

\begin{table}
\centering
\renewcommand{\arraystretch}{0.9} 
\caption{Mean number of agents choosing each route by game. Data is aggregated across rounds and trials as described in Section \ref{sec:metrics}. Standard errors are omitted from \cite{rapoport_choice_2009} data because they were not computed across trials and were thus not comparable.}
\label{tab:experiment1}
\begin{adjustbox}{max width=\textwidth}
\begin{tabular}{@{}lccccccc@{}}
\toprule
\textbf{Game} & \multicolumn{2}{c}{\textbf{Game A (two routes)}} & \multicolumn{3}{c}{\textbf{Game B (three routes)}} \\
\cmidrule(r){2-3} \cmidrule(l){4-6}
 & \textbf{(O--L--D)} & \textbf{(O--R--D)} & \textbf{(O--L--D)} & \textbf{(O--R--D)} & \textbf{(O--L--R--D)} \\
\midrule
\rowcolor[HTML]{F3F3F3} F-PE & 9.03 & 8.97 & 6.38 & 6.88 & 4.74 \\
\rowcolor[HTML]{F3F3F3} & (7.26) & (7.26) & (6.59) & (6.88) & (2.53) \\
F-PO & 8.97 & 9.03 & 4.50 & 5.25 & 8.25 \\
& (7.24) & (7.24) & (3.79) & (4.56) & (3.44) \\
\rowcolor[HTML]{F3F3F3} F-RE & 8.98 & 9.02 & 3.94 & 4.16 & 9.90 \\
\rowcolor[HTML]{F3F3F3} & (6.71) & (6.71) & (4.07) & (4.14) & (1.48) \\
F-RO & 8.79 & 9.21 & 0.33 & 0.57 & 17.52 \\
& (7.99) & (7.99) & (0.52) & (0.66) & (0.66) \\
\rowcolor[HTML]{F3F3F3} S-PE & 8.90 & 9.10 & 3.47 & 3.65 & 10.88 \\
\rowcolor[HTML]{F3F3F3} & (3.52) & (3.52) & (1.99) & (1.92) & (2.46) \\
S-PO & 9.35 & 8.65 & 1.07 & 1.02 & 15.90 \\
& (2.27) & (2.27) & (1.65) & (1.64) & (2.77) \\
\rowcolor[HTML]{F3F3F3} S-RE & 8.75 & 9.25 & 0.24 & 0.39 & 17.36 \\
\rowcolor[HTML]{F3F3F3} & (2.94) & (2.94) & (0.82) & (1.49) & (1.93) \\
S-RO & 7.86 & 10.14 & 0.11 & 0.17 & 17.71 \\
& (2.11) & (2.11) & (0.49) & (0.95) & (1.26) \\
\rowcolor[HTML]{F3F3F3} EXP3 & 9.01 & 8.99 & 2.11 & 2.08 & 13.81 \\
\rowcolor[HTML]{F3F3F3} & (0.01) & (0.01) & (0.02) & (0.02) & (0.03) \\
MWU & 9.01 & 8.99 & 2.08 & 2.05 & 13.87 \\
& (0.01) & (0.01) & (0.02) & (0.02) & (0.02) \\
\rowcolor[HTML]{F3F3F3} Human subjects & 9.02 & 8.98 & 1.72 & 1.47 & 14.82 \\
\rowcolor[HTML]{F3F3F3} \text{\cite{rapoport_choice_2009}} &  &  &  &  &  \\
Pure-strategy equilibrium & 9.00 & 9.00 & 0.00 & 0.00 & 18.00 \\
&  &  &  &  &  \\
\bottomrule
\end{tabular}
\end{adjustbox}
\renewcommand{\arraystretch}{1} 
\Description{}
\end{table}

\begin{figure}[ht]
    \centering
    \begin{subfigure}{0.48\textwidth}
        \centering
        \includegraphics[width=\linewidth]{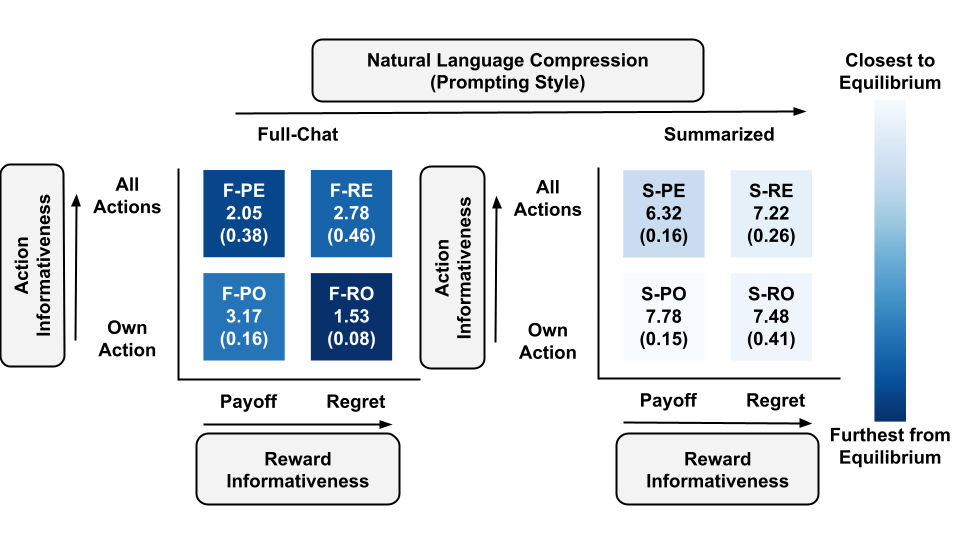}
        \caption{Fig.~\ref{fig:agg_routes-A}. Mean agent route choices in Game A. We measure the number of agents on the least-congested route in each round.}
        \label{fig:agg_routes-A}
    \end{subfigure}
    \hfill
    \begin{subfigure}{0.48\textwidth}
        \centering
        \includegraphics[width=\linewidth]{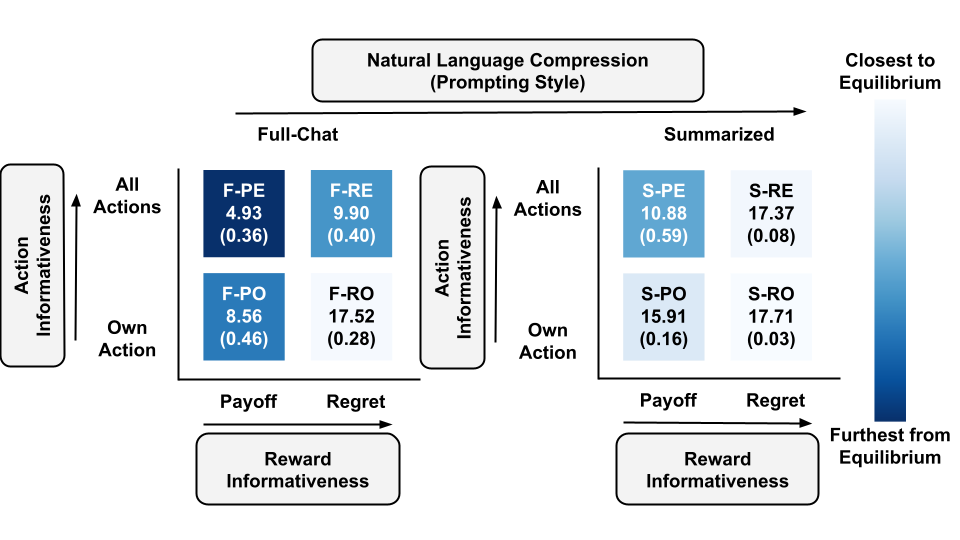}
        \caption{Fig.~\ref{fig:agg_routes-B}. Mean agent route choices in Game B. We measure the number of agents on the dominant route.}
        \label{fig:agg_routes-B}
    \end{subfigure}
    \label{fig:agg_routes}
    \Description{}
\end{figure}

\begin{figure}[ht]
    \centering
    \begin{subfigure}{0.48\textwidth}
        \centering
        \includegraphics[width=\linewidth]{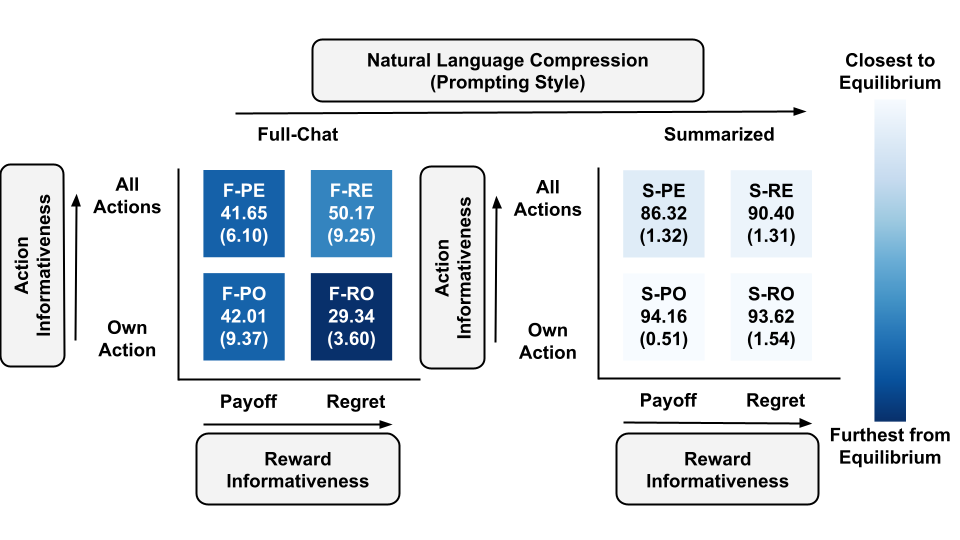}
        \caption{Fig.~\ref{fig:agg_payoffs-A}. Mean agent payoffs in Game A.}
        \vspace{0.8cm}
        \label{fig:agg_payoffs-A}
    \end{subfigure}
    \hfill
    \begin{subfigure}{0.48\textwidth}
        \centering
        \includegraphics[width=\linewidth]{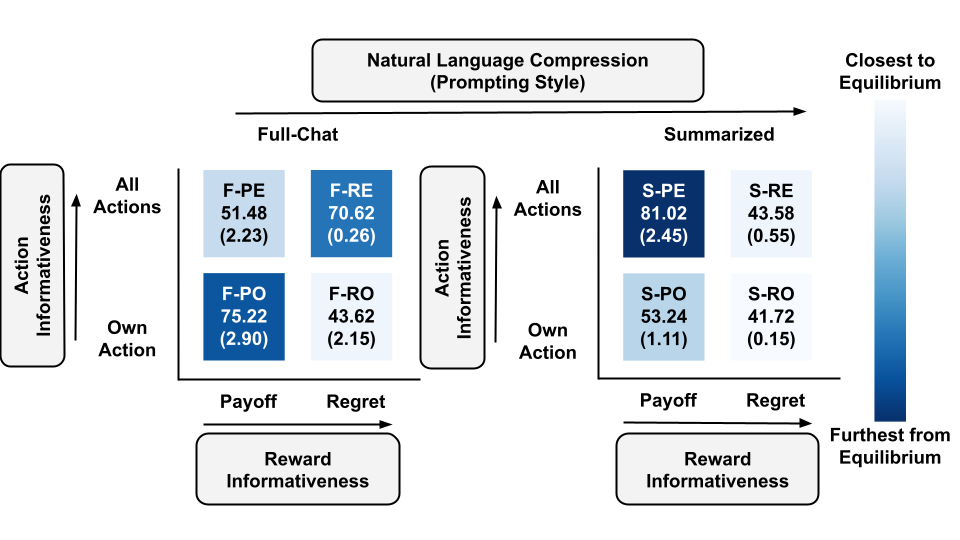}
        \caption{Fig.~\ref{fig:agg_payoffs-B}. Mean agent payoffs in Game B. Color axis is reversed to indicate that lower payoffs are more desirable.}
        \label{fig:agg_payoffs-B}
    \end{subfigure}
    \label{fig:agg_payoffs}
    \Description{}
\end{figure}

\begin{figure}[ht]
    \centering
    \begin{subfigure}{0.48\textwidth}
        \centering
        \includegraphics[width=\linewidth]{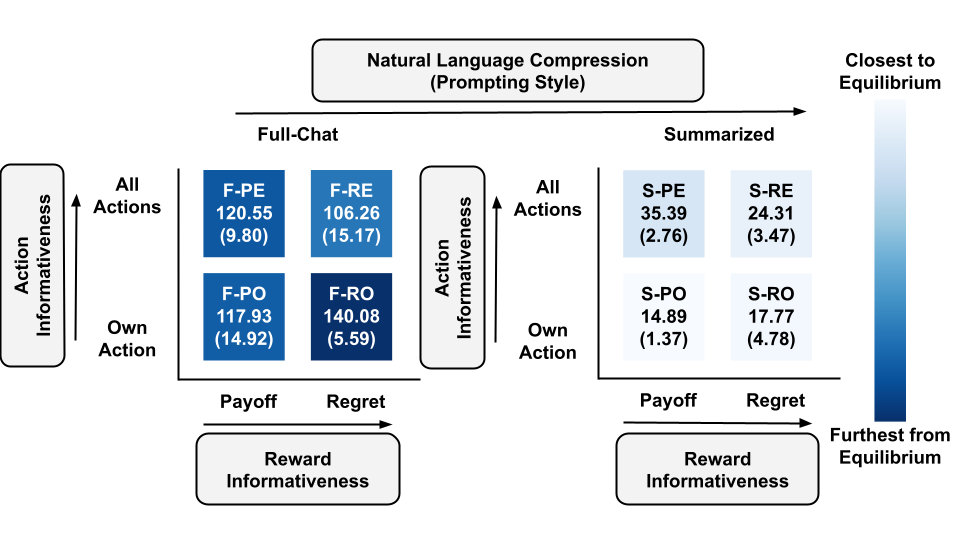}
        \caption{Fig.~\ref{fig:agg_regrets-A}. Mean agent regrets in Game A.}
        \label{fig:agg_regrets-A}
    \end{subfigure}
    \hfill
    \begin{subfigure}{0.48\textwidth}
        \centering
        \includegraphics[width=\linewidth]{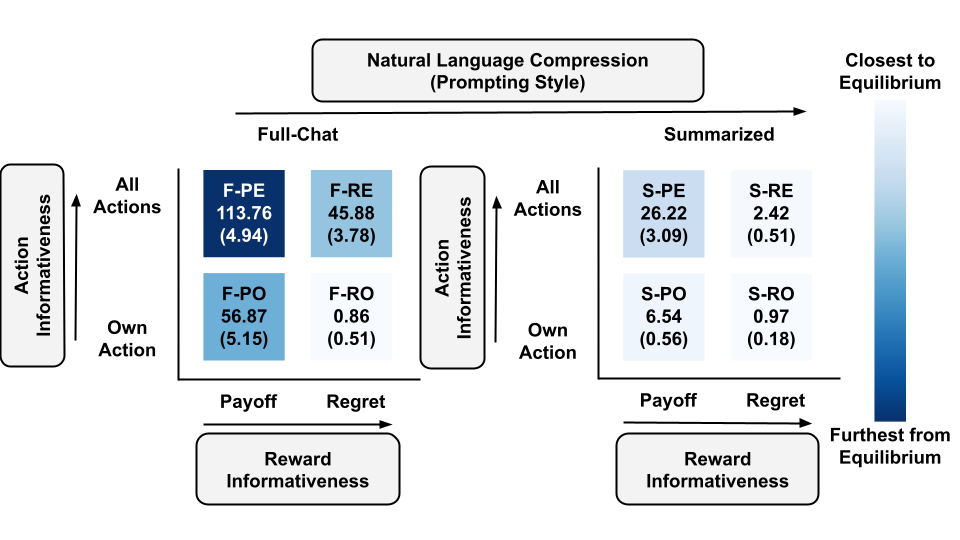}
        \caption{Fig.~\ref{fig:agg_regrets-B}. Mean agent regrets in Game B.}
        \label{fig:agg_regrets-B}
    \end{subfigure}
    \label{fig:agg_regrets}
    \Description{Mean agent regrets in Game B.}
\end{figure}

\begin{figure}[ht]
    \centering
    \begin{subfigure}{0.48\textwidth}
        \centering
        \includegraphics[width=\linewidth]{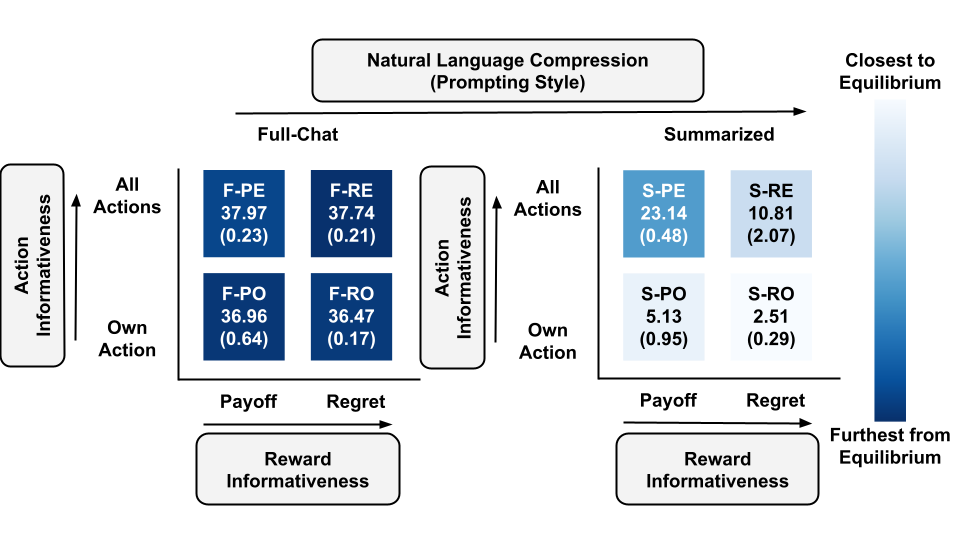}
        \caption{Fig.~\ref{fig:agg_switches-A}. Mean agent switch count in Game A.}
        \label{fig:agg_switches-A}
    \end{subfigure}
    \hfill
    \begin{subfigure}{0.48\textwidth}
        \centering
        \includegraphics[width=\linewidth]{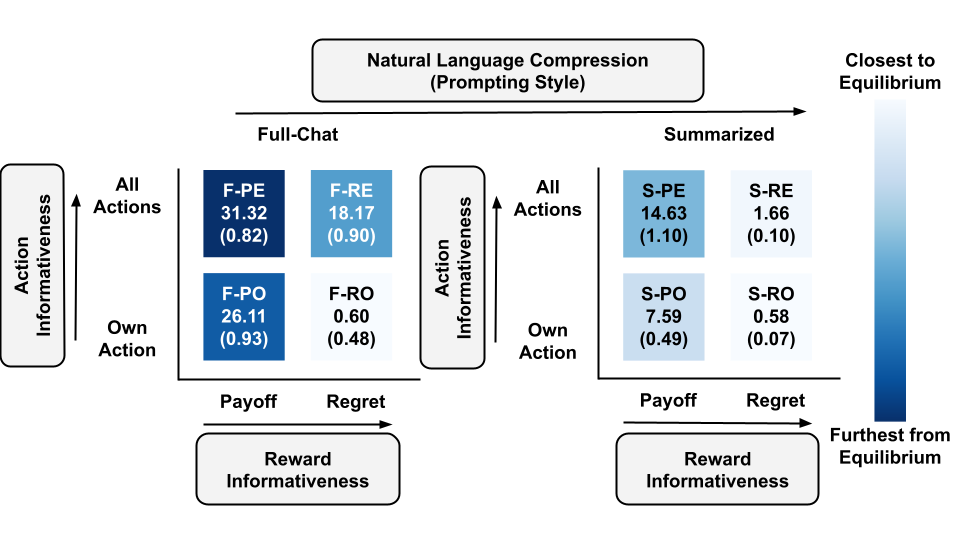}
        \caption{Fig.~\ref{fig:agg_switches-B}. Mean agent switch count in Game B.}
        \label{fig:agg_switches-B}
    \end{subfigure}
    \label{fig:agg_switches}
    \Description{}
\end{figure}

Table \ref{tab:experiment1} displays the mean number of agents on all routes in both games. For comparison, the table also includes data from \cite{rapoport_choice_2009}; results from two learning algorithms, Multiplicative Weights Update (MWU) \cite{freund_adaptive_1999} and EXP3 \cite{auer_nonstochastic_2002}; and pure strategy Nash equilibrium route choices. In Figures \ref{fig:agg_routes-A}-\ref{fig:agg_switches-B}, we present our findings for the aggregate metrics described in Section \ref{sec:metrics}.  Each panel presents results organized by the three axes that define our representations.  For each game, we present full-chat prompting style results on the left and summarized prompting style results on the right.  In each $2x2$ matrix, the rows represent action informativeness, and the columns represent reward informativeness.  In each graphic, lighter colors represent mean (aggregate) outcomes that are closer to equilibrium behavior.

Throughout our analysis of the results, our main emphasis is to understand which choices of representation lead agents to better replicate equilibrium behavior.  Below we describe our most salient findings.

\paragraph{The summarized prompting style leads agents to more closely replicate equilibrium behavior than the full prompting style.}  This is the clearest finding in our results, as evidenced by comparing the left and right panels in each subfigure across both Games A and B.  As measured by all metrics, with few exceptions, the summarized prompting style leads to behavior that is closer to equilibrium (lighter color in the figure) than the full-chat prompting style.

Some specific examples are particularly informative.  When considering summarized representations in Game A, the mean number of agents on a focal route (Figure \ref{fig:agg_routes-A}) is approximately 40\% for summarized representations, and closer to 15\% for full-chat representations (in equilibrium 50\% of agents should choose the focal route). Summarized prompting also leads to much higher payoffs in Game A, and generally lower payoffs in Game B, relative to full chat prompting (see Figures \ref{fig:agg_payoffs-A}-\ref{fig:agg_payoffs-B}).  Summarized representations lead to significantly lower regret, as seen in Figures \ref{fig:agg_regrets-A}-\ref{fig:agg_regrets-B}---regrets are 4-5x lower using summarized representations in Game A, and regret is near zero using summarized representations in Game B.  Finally, and interestingly, summarized representations lead to more stable behavior, as measured by mean number of switches per agent (Figures \ref{fig:agg_switches-A}-\ref{fig:agg_switches-B}).  As we discuss in Section \ref{sec:discussion}, one likely reason for this behavior is that full-chat representations induce myopia in the LLM agents in our simulations.

\paragraph{When there are distinctions in the value of actions, regret-based representations lead agents closer to equilibrium behavior than payoff-based representations.}  To elucidate this finding, we compare the left and right columns of each 2x2 submatrix in our figures.  Note that a representation that reports regret (rather than payoff) will have the greatest salience in Game B, because in this game there is a weak dominant strategy (the route \textit{O-L-R-D}), {\em and} this strategy yields lower payoffs. By contrast, in Game A, both routes are equally attractive in equilibrium, {\em and} both yield high payoffs.  

In our findings, we see therefore that the impact of regret-based vs.~payoff-based representations is not as stark for Game A.  However, for Game B, we see that there is a strong impact of this choice of representation.  In Figure \ref{fig:agg_routes-B}, we generally see a larger number of agents choosing the weak dominant strategy in regret-based representations, and a lower standard error as well, relative to payoff-based representations (the only exception is F-PE and F-RE, which have nearly identical standard errors, suggesting that regret may not reduce variance as strongly among full-chat representations).  A similar behavior is found when comparing mean payoffs as well (Figure \ref{fig:agg_payoffs-B}).  The effects of regret-based representations are even more dramatic in Game B when we study mean regret (Figure \ref{fig:agg_regrets-B}) and mean switching behavior (Figure \ref{fig:agg_switches-B}).  Taken together, these findings make clear that when there is useful ``gradient'' information comparing the chosen action to the optimal action, LLM agents benefit from being provided explicitly with this information in the form of regret.

By contrast, in the results for Game A across all figures, the results are more mixed, consistent with the observation that in this game either route is equally preferable.  Notably, switching behavior is not necessarily reduced in Game A (Figure \ref{fig:agg_switches-A}) when regrets are provided, emphasizing the much stronger role that summarization plays in this regard.

\paragraph{Seeing only one's own action tends to reduce instability in agent behavior.}  Here we focus specifically on Figures \ref{fig:agg_switches-A}-\ref{fig:agg_switches-B}.  Notice that for both Games A {\em and} B, we see lower switching behavior with lower action informativenes.  Lower action informativeness here helps reduce variability in agents' context, and thus their subsequent behavior.  Indeed, this is consistent with our discussion of summarization above.

Interestingly, for the other metrics, we also see some evidence of the potential benefits of lower action informativeness, particularly in Game B: the mean number of agents on a focal route (Figure \ref{fig:agg_routes-B}) and mean regret (Figure \ref{fig:agg_regrets-B}) are closer to equilibrium behavior.  In general, these findings suggest that to the extent that action informativeness helps guide agents closer to equilibrium behavior, it is more likely to do so by {\em lowering} (rather than increasing) the information provided about others' actions.

\subsection{Sample Path Behavior}
\label{sec:dynamics}

\begin{figure}[htbp]
    \centering
    \begin{subfigure}[b]{0.48\textwidth}
        \centering
        \includegraphics[width=\textwidth]{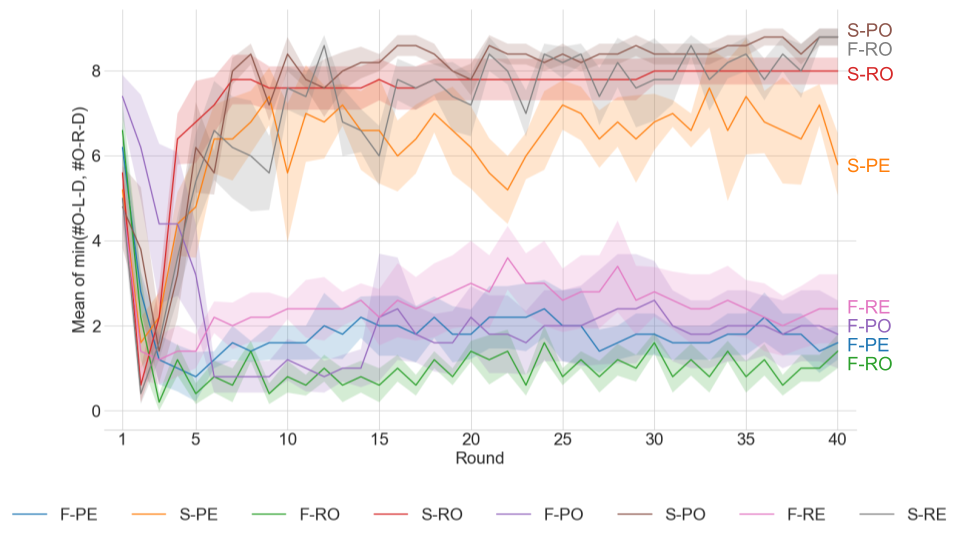}
        \caption{Fig. \ref{fig:LLM-route-trends-A}. Round-to-round mean agent route choices in Game A (number of agents on the least-congested route).}
        \label{fig:LLM-route-trends-A}
        \Description{}
    \end{subfigure}
    \hfill
    \begin{subfigure}[b]{0.48\textwidth}
        \centering
        \includegraphics[width=\textwidth]{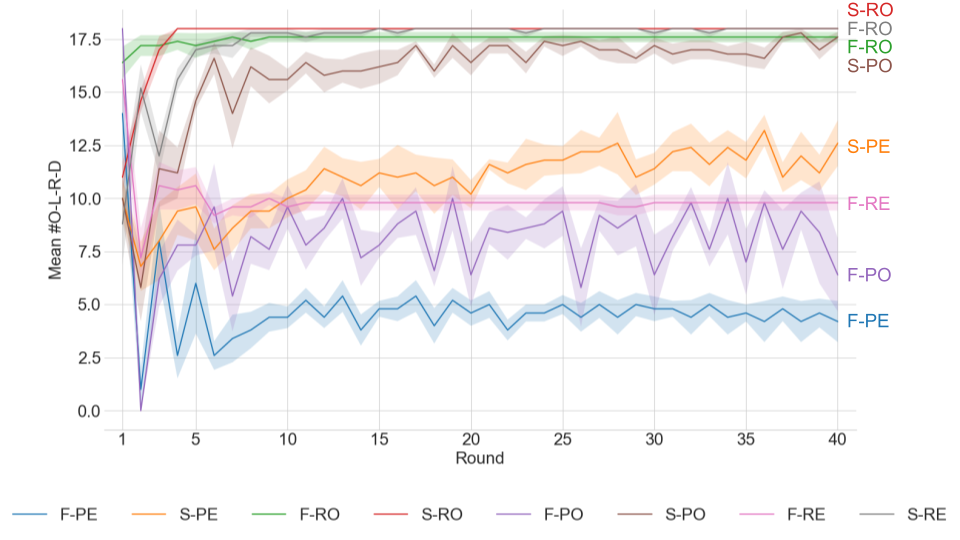}
        \caption{Fig. \ref{fig:LLM-route-trends-B}. Round-to-round mean agent route choices in Game B (number of agents on the dominant route).}
        \label{fig:LLM-route-trends-B}
        \Description{}
    \end{subfigure}
    \label{fig:LLM-route-trends}
\end{figure}

\begin{figure}[htbp]
    \centering
    \begin{subfigure}[b]{0.48\textwidth}
        \centering
        \includegraphics[width=\textwidth]{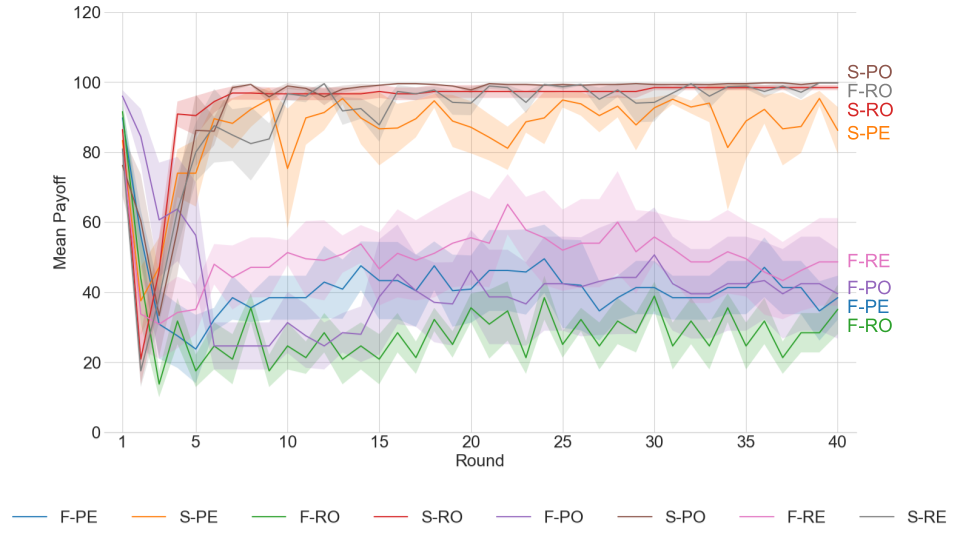}
        \caption{Fig. \ref{fig:LLM-reward-trends-A}. Round-to-round mean agent reward in Game A.}
        \label{fig:LLM-reward-trends-A}
        \Description{}
    \end{subfigure}
    \hfill
    \begin{subfigure}[b]{0.48\textwidth}
        \centering
        \includegraphics[width=\textwidth]{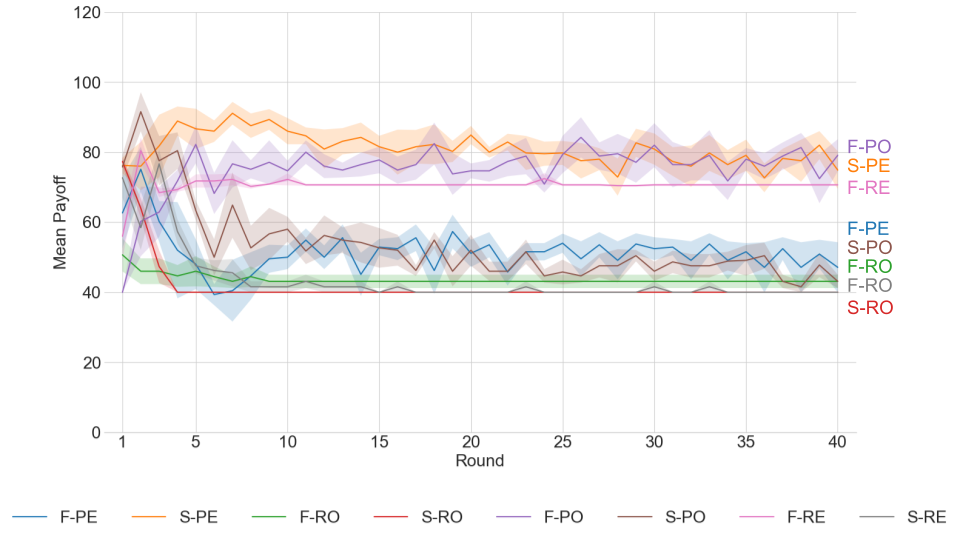}
        \caption{Fig. \ref{fig:LLM-reward-trends-B}. Round-to-round mean agent reward in Game B.}
        \label{fig:LLM-reward-trends-B}
        \Description{}
    \end{subfigure}
    \label{fig:LLM-reward-trends}
\end{figure}

\begin{figure}[htbp]
    \centering
    \begin{subfigure}[b]{0.48\textwidth}
        \centering
        \includegraphics[width=\textwidth]{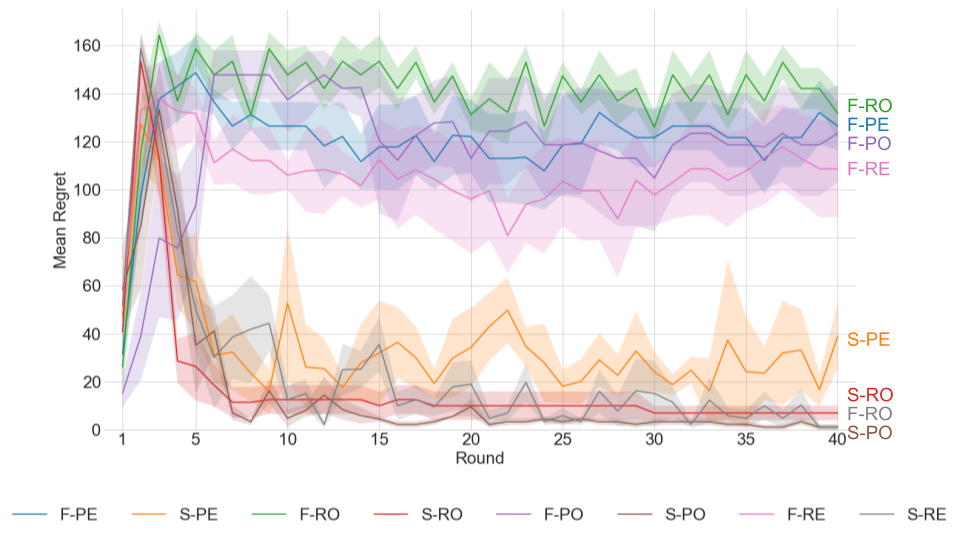}
        \caption{Fig. \ref{fig:LLM-regret-trends-A}. Round-to-round mean agent regret in Game A.}
        \label{fig:LLM-regret-trends-A}
        \Description{}
    \end{subfigure}
    \hfill
    \begin{subfigure}[b]{0.48\textwidth}
        \centering
        \includegraphics[width=\textwidth]{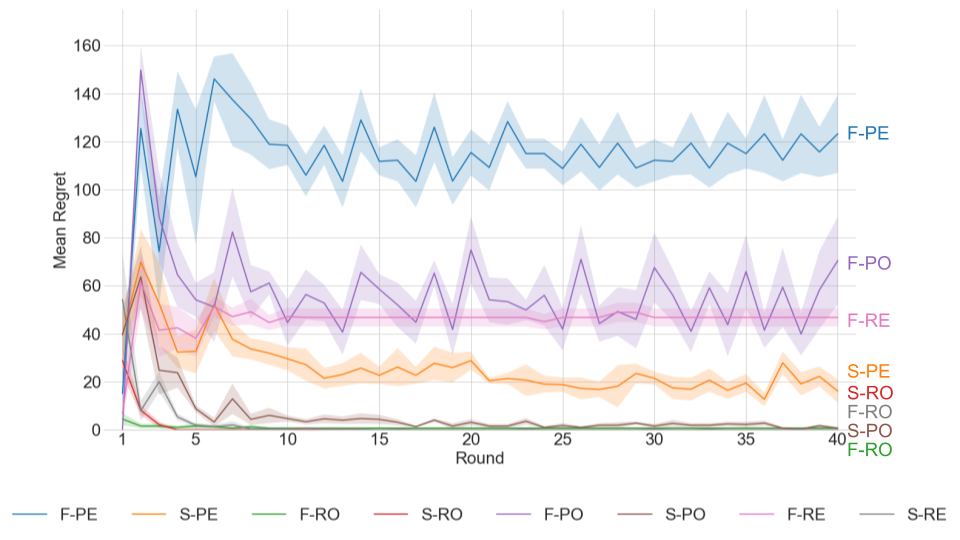}
        \caption{Fig. \ref{fig:LLM-regret-trends-B}. Round-to-round mean agent regret in Game B.}
        \label{fig:LLM-regret-trends-B}
        \Description{}
    \end{subfigure}
    \label{fig:LLM-regret-trends}
\end{figure}

\begin{figure}[htbp]
    \centering
    \begin{subfigure}[b]{0.48\textwidth}
        \centering
        \includegraphics[width=\textwidth]{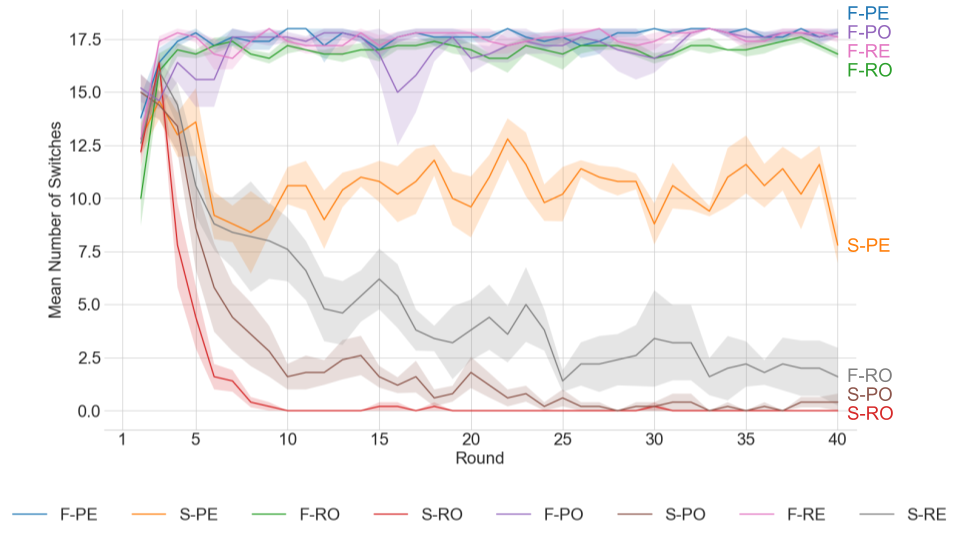}
        \caption{Fig. \ref{fig:LLM-switch-trends-A}. Round-to-round mean agent switch count in Game A.}
        \label{fig:LLM-switch-trends-A}
        \Description{}
    \end{subfigure}
    \hfill
    \begin{subfigure}[b]{0.48\textwidth}
        \centering
        \includegraphics[width=\textwidth]{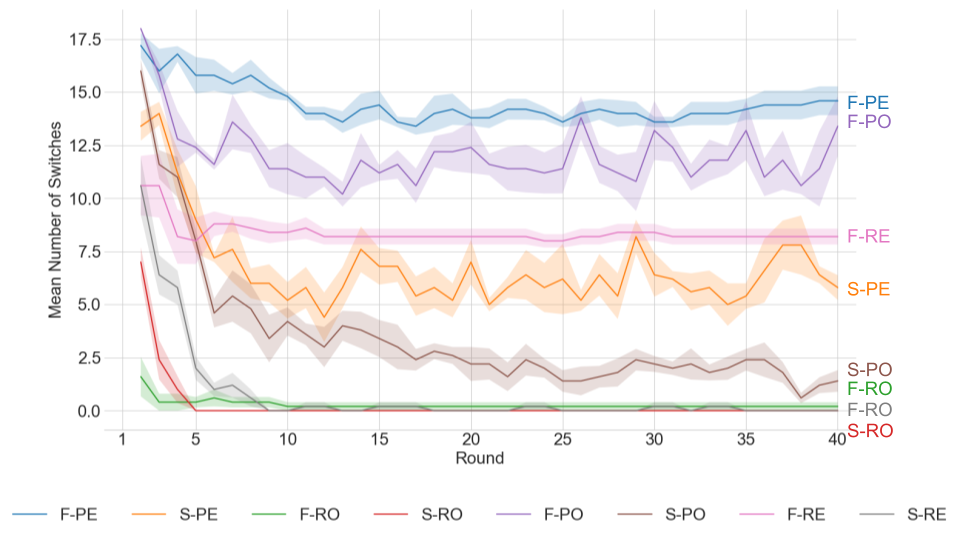}
        \caption{Fig. \ref{fig:LLM-switch-trends-B}. Round-to-round mean agent switch count in Game B.}
        \label{fig:LLM-switch-trends-B}
        \Description{}
    \end{subfigure}
    \label{fig:LLM-switch-trends}
\end{figure}

To expand on our findings above, in this section we analyze the dynamic behavior of agent play by looking at mean sample paths over the 40 round horizon. Figures \ref{fig:LLM-route-trends-A}-\ref{fig:LLM-switch-trends-B} display the round-by-round behavior of each of our four key metrics over time.  

These graphs amplify our first finding above, namely, that summarized prompting generally leads to behavior that is closer to equilibrium in both games, both in terms of speed of convergence, and in terms of stability of behavior.  In all graphs, there is generally a clear separation between the summarized representations and the full-chat representations.  Perhaps the most striking visual finding is the distinction between round-to-round regret behavior between representations, displayed in Figures \ref{fig:LLM-regret-trends-A} and \ref{fig:LLM-regret-trends-B} for Games A and B, respectively. Here, summarized prompting dramatically reduces regret in both games to a similar degree.  As previously observed, summarized prompting generally leads to less switching behavior over time than with full-chat prompting (Figures \ref{fig:LLM-switch-trends-A}-\ref{fig:LLM-switch-trends-B}).  

There is some nuance related to our second finding; namely, the potential impact of regret-based representations over payoff-based representations in guiding agents towards equilibrium actions.  For example, in Game B, we see that F-RO (a regret-based representation) actually exhibits reasonably fast convergence to equilibrium (cf.~Figure \ref{fig:LLM-route-trends-B}), and also lower regret over time (cf.~Figure \ref{fig:LLM-regret-trends-B}).  These effects suggest that the regret-based representation offsets some of the complexity of the full chat prompting style in providing useful context to the LLM agent---since, again, in Game B, the primary goal is to identify the weak dominant strategy \textit{O-L-R-D}.  We also note here that this type of visualization reveals the potential for intriguing interactions {\em between} the different representation axes we have investigated.

Finally, by examining Figures \ref{fig:LLM-switch-trends-A}-\ref{fig:LLM-switch-trends-B}, we can see the impact of action informativeness on behavior over time. In particular, conditional on using prompting style, not providing other agents' actions generally reduces the decision instability of agents in a persistent manner over the time horizon.  


\subsection{Convergence to Equilibrium}

\begin{figure}
    \centering
    \begin{subfigure}{0.48\textwidth}
        \centering
        \includegraphics[width=\linewidth]{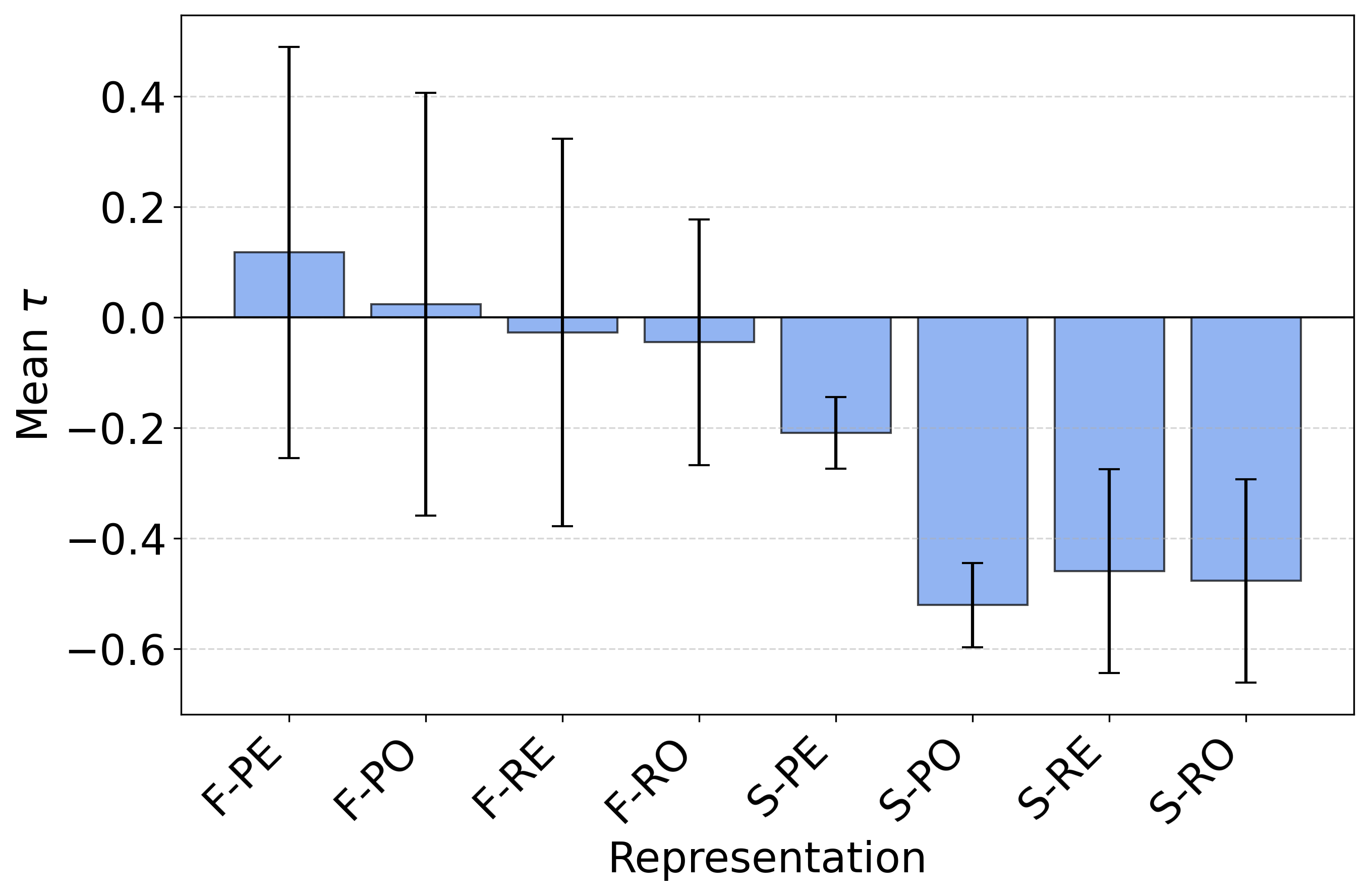}
        \caption{Fig.~\ref{fig:LLM-tau-A}. Mean $\tau$ across trials, Game A}
        \label{fig:LLM-tau-A}
    \end{subfigure}
    \hfill
    \begin{subfigure}{0.48\textwidth}
        \centering
        \includegraphics[width=\linewidth]{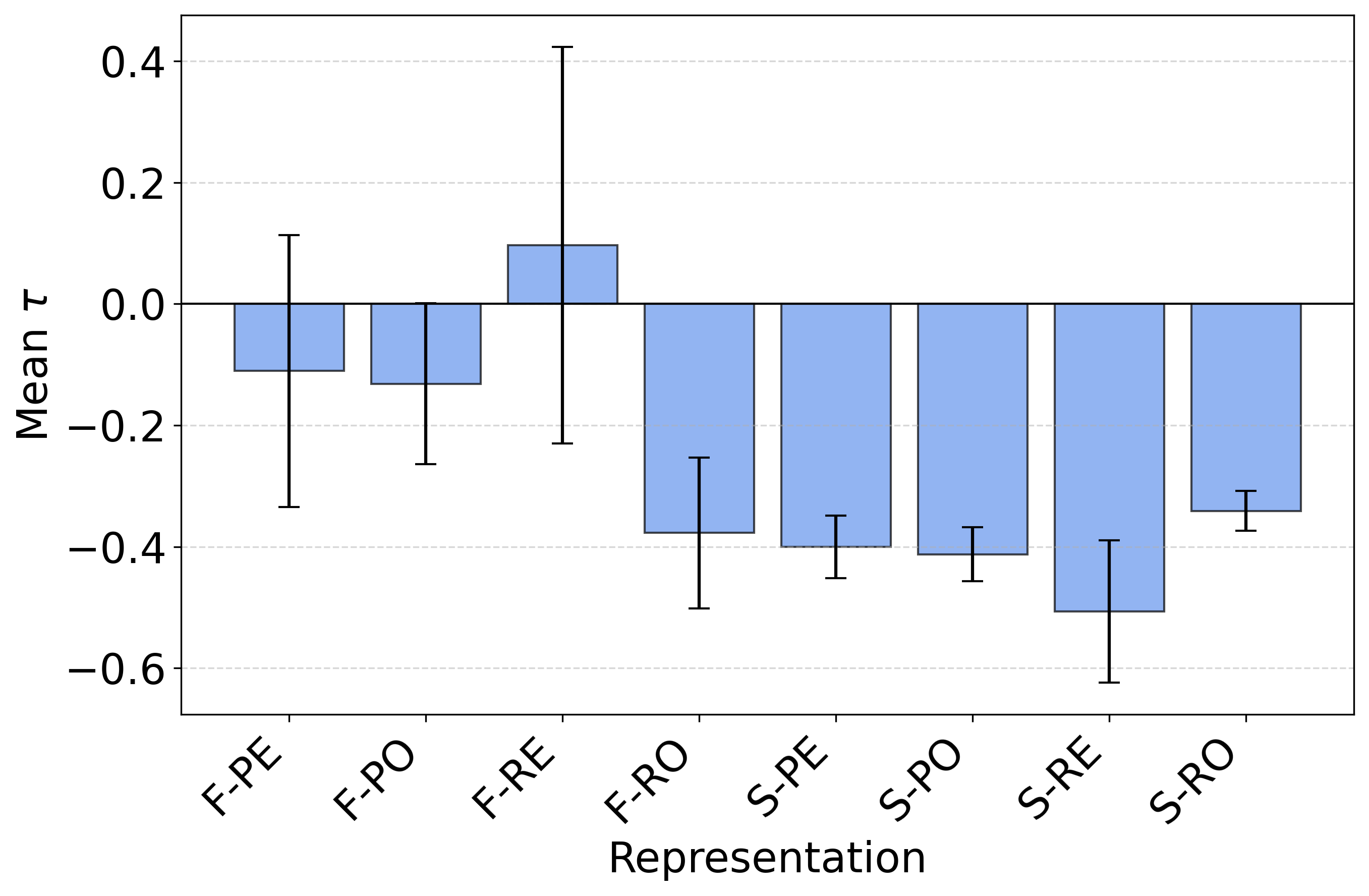}
        \caption{Fig.~\ref{fig:LLM-tau-B}. Mean $\tau$ across trials, Game B}
        \label{fig:LLM-tau-B}
    \end{subfigure}
    \caption{Kendall's $\tau$ between round number and equilibrium deviation score in Game A and Game B.}
    \label{fig:LLM-tau}
    \Description{}
\end{figure}

Finally, we study convergence to equilibrium via Kendall's $\tau$ between round number and equilibrium deviation score in Game A (Fig. \ref{fig:LLM-tau-A}) and Game B (Fig. \ref{fig:LLM-tau-B}), cf.~Section \ref{sec:metrics}. Again, a negative (resp., positive) $\tau$ indicates agents are approaching (resp., diverging from) equilibrium as rounds progress; $\tau\approx0$ suggests no systematic convergence trend.  We note that we were unable to compute $\tau$ for two trials of representation F-RO in Game B; these trials exhibited constant deviation scores, and Kendall's $\tau$ is undefined if there is no variability in one of the variables.

In Game A, summarized representations (S-PE, S-RE, S-PO, S-RO) exhibit negative $\tau$ values, indicating steady decreases in deviation scores (likewise convergence to equilibrium). In contrast, full-chat representations (F-PE, F-RE, F-PO, F-RO) show weak or no convergence, with $\tau$ values centered around zero and high variance among trials. Notably, F-RO achieves the best equilibrium convergence among full-chat representations, reinforcing that regret-based feedback encourages convergence.

In Game B, summarized representations yield the most negative $\tau$ values. Regret-based representations (F-RO, S-RE, S-RO) also show significantly negative $\tau$, suggesting that regret encourages consistent learning. These findings quantify our temporal observations from sample paths in the preceding section.

\subsection{Comparison with Learning Algorithms}

For completeness, we also compare LLM agent behavior to two well-known no-regret learning algorithms: Multiplicative Weights Update (MWU) \cite{freund_adaptive_1999} and EXP3 \cite{auer_nonstochastic_2002}. 
MWU operates under {\em full} feedback, meaning it has access to the payoffs of \textit{all} possible actions in each round, even those it did not take. In contrast, EXP3 operates under {\em bandit} feedback, meaning it only observes the payoff of the action it actually took. The precise mathematical formulation of each algorithm is available in Appendix \ref{sec:rl_algs}. Both were simulated under identical conditions to the LLM agents ($n = 18$ agents, $T = 40$ rounds) over 50 trials. MWU used a learning rate of $0.75$, and EXP3 an exploration rate of $0.75$. Payoffs were rescaled from $[0,400]$ to $[0,1]$ to ensure convergence within 40 rounds.

\begin{figure}[ht]
    \centering
    \begin{tabular}{@{}c@{\quad}c@{}}
        \begin{subfigure}[b]{0.35\textwidth}
            \centering
            \includegraphics[width=\linewidth]{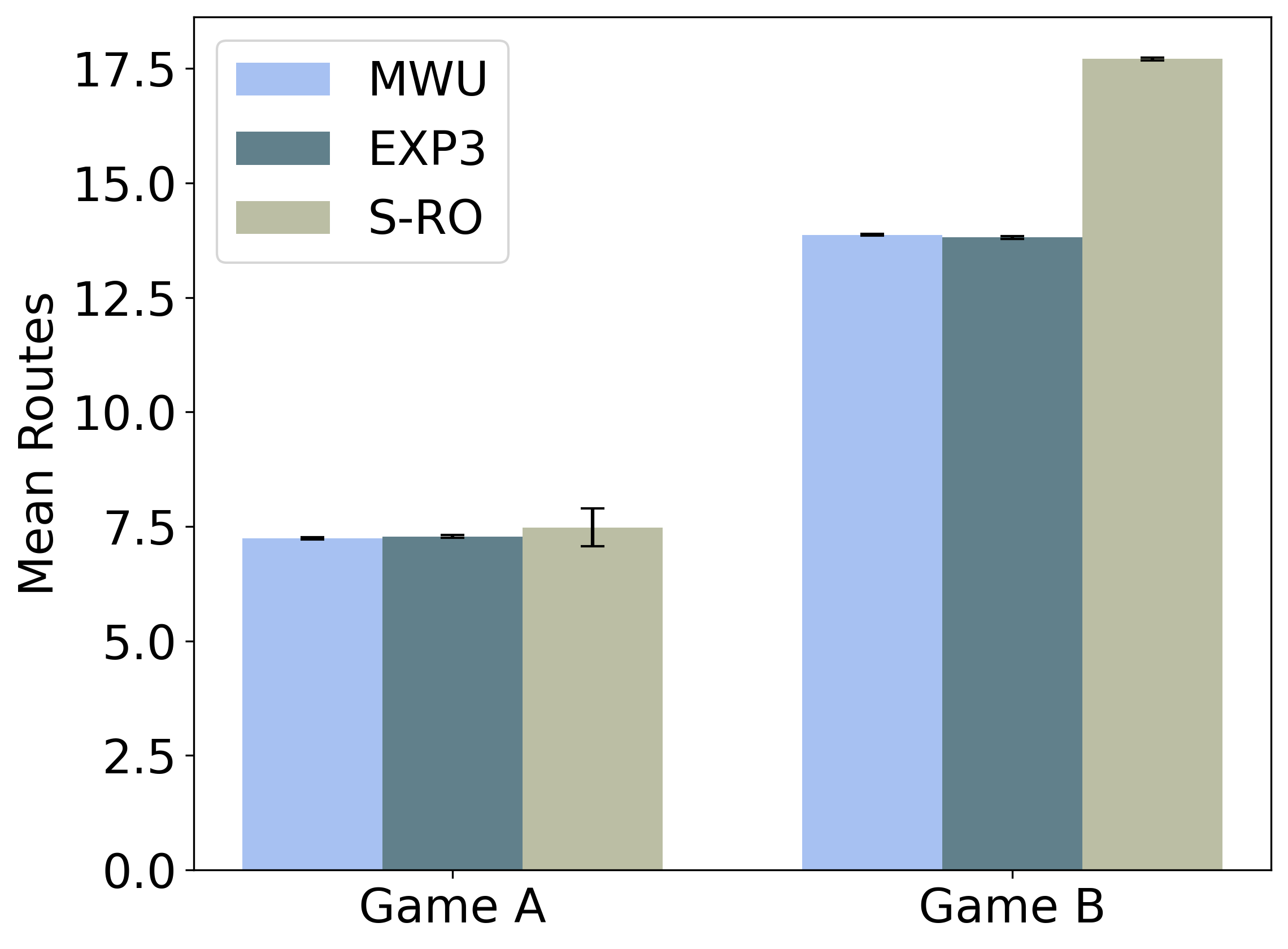}
            \caption{Fig. \ref{fig:mean_routes}. Mean agent route choices.}
            \label{fig:mean_routes}
        \end{subfigure} &
        \begin{subfigure}[b]{0.35\textwidth}
            \centering
            \includegraphics[width=\linewidth]{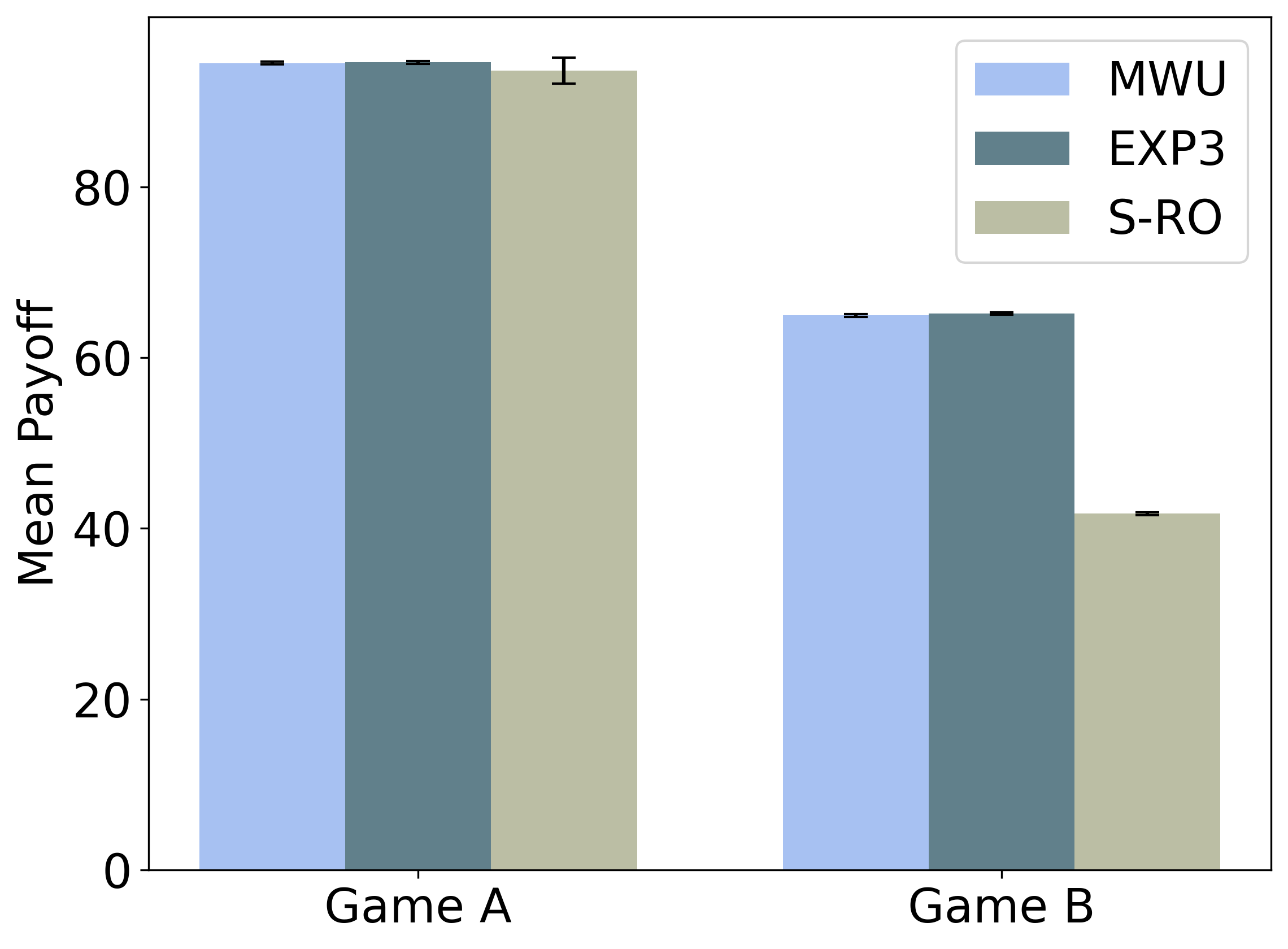}
            \caption{Fig. \ref{fig:mean_payoff}. Mean agent payoff.}
            \label{fig:mean_payoff}
        \end{subfigure} \\[1ex] 
        \begin{subfigure}[b]{0.35\textwidth}
            \centering
            \includegraphics[width=\linewidth]{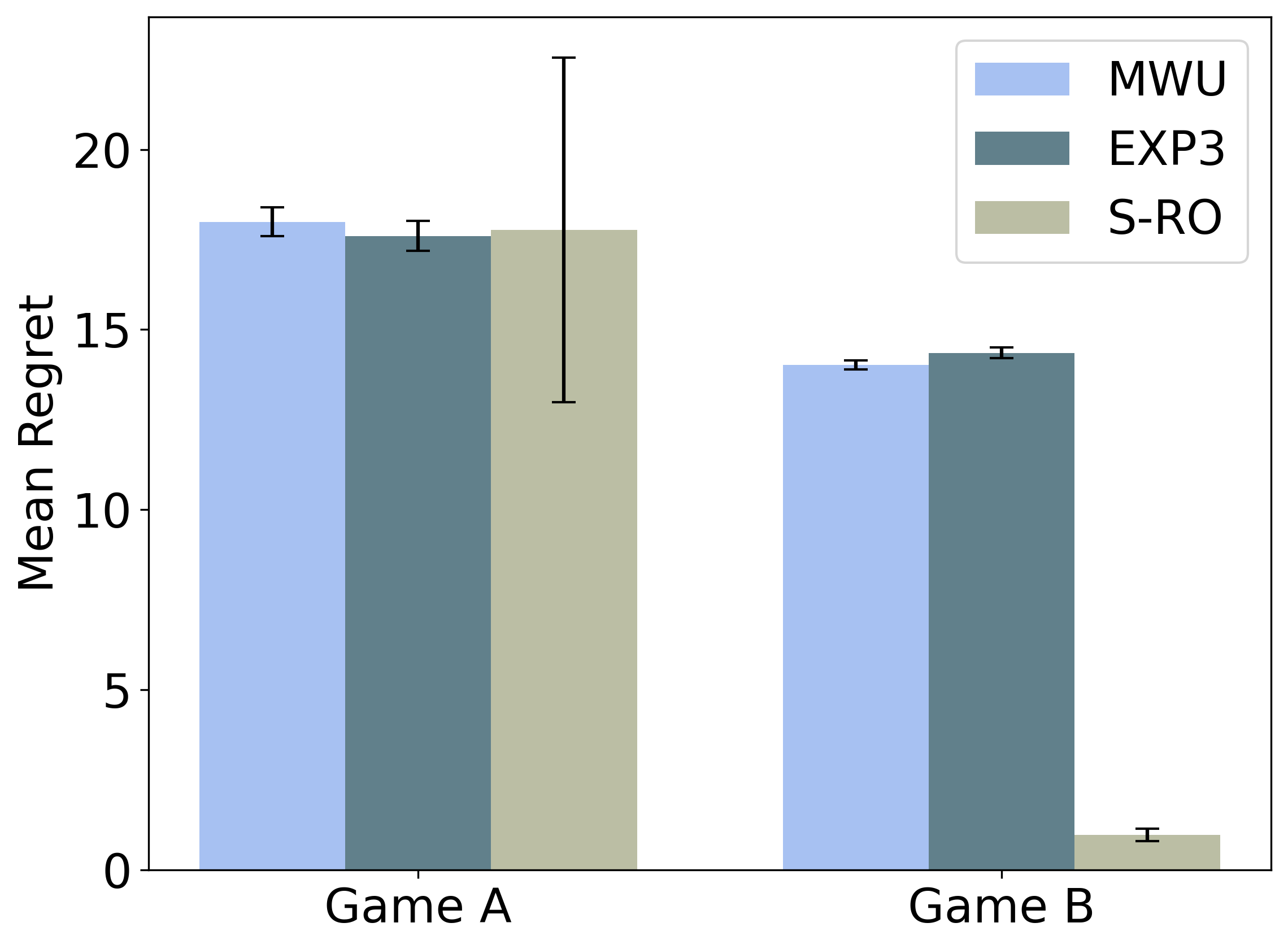}
            \caption{Fig. \ref{fig:mean_regret}. Mean agent regret.}
            \label{fig:mean_regret}
        \end{subfigure} &
        \begin{subfigure}[b]{0.35\textwidth}
            \centering
            \includegraphics[width=\linewidth]{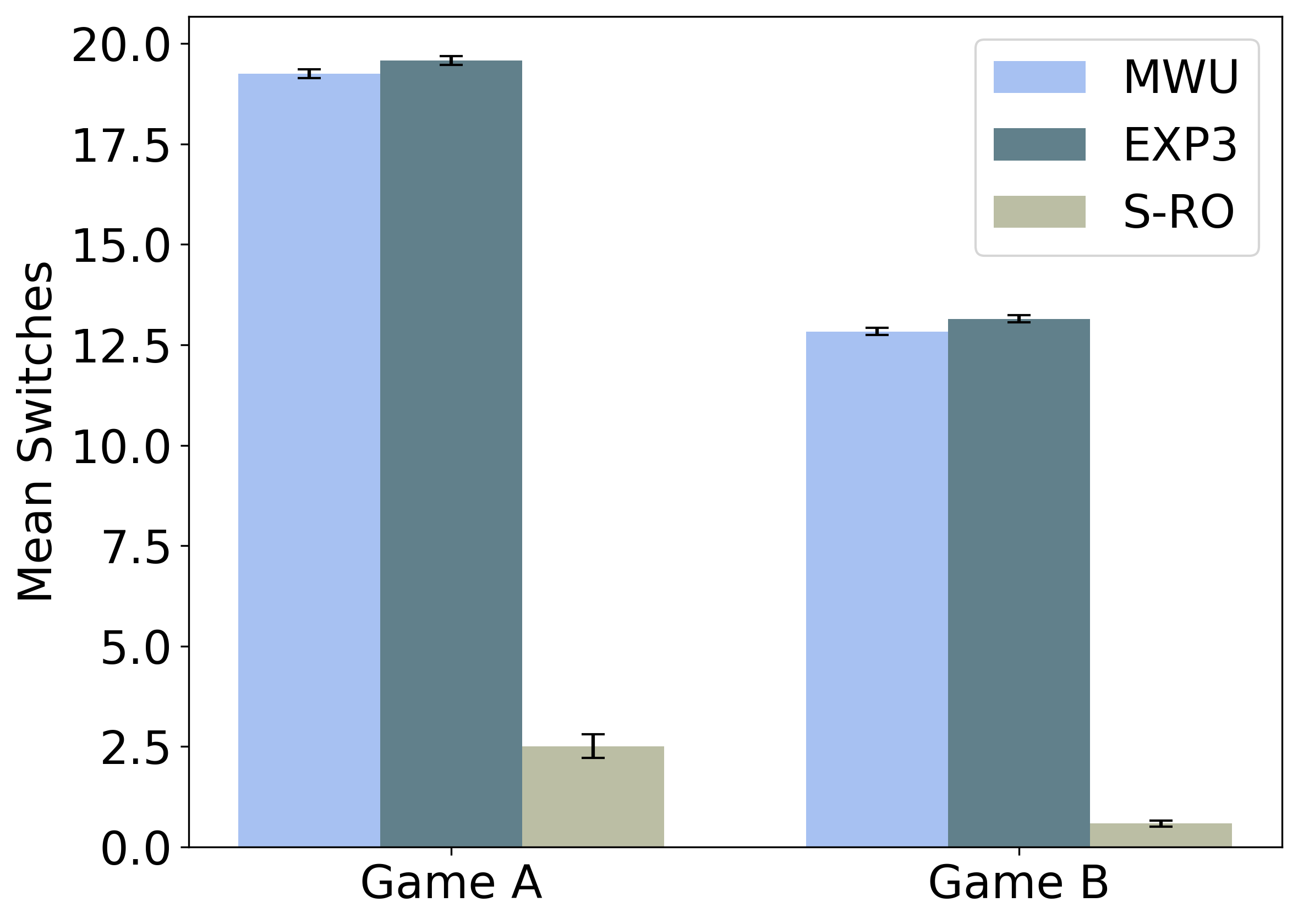}
            \caption{Fig. \ref{fig:mean_switches}. Mean agent switch count.}
            \label{fig:mean_switches}
        \end{subfigure}
    \end{tabular}
    \caption{Comparison of LLM agent performance under S-RO with learning algorithms EXP3 and MWU across four aggregate metrics in Games A and B.}
    \label{fig:comparison}
    \Description{}
\end{figure}

Figure~\ref{fig:comparison} compares the performance of EXP3 and MWU against LLM agent performance under S-RO---the representation that resulted in LLM behavior closest to equilibrium---across our four aggregate metrics from Section~\ref{sec:agg-metrics}. In Figures~\ref{fig:mean_routes}-\ref{fig:mean_switches}, S-RO exhibits behavior closer to theoretical equilibrium than the two learning algorithms on all four aggregate metrics in Game B. In Game A as well, we observe that while mean routes, payoffs, and regrets are similar across S-RO, MWU, and EXP3 (Figures~\ref{fig:mean_routes}-\ref{fig:mean_regret}), S-RO results in far fewer switches on average (Figure~\ref{fig:mean_switches}), suggesting more stability at equilibrium under S-RO.

\subsection{Discussion and Future Directions}
\label{sec:discussion}

Our paper makes two primary contributions.  First, we introduce a unifying framework for constructing natural language state representations in repeated games, systematically characterizing them along the axes of action informativeness, reward informativeness, and prompting style (or natural language compression). This framework can be used by researchers who seek to understand and empirically test the relationship between various axes of natural language state informativeness on LLM agent behavior in their own game settings. Beyond dynamic games, this framework can be used for the augmentation and optimization of LLM agent as decision-makers in dynamic settings.

Second, we apply this framework to a repeated selfish routing game to rigorously evaluate how the dimensions of state representation influence LLM decision-making, and in particular use several evaluation metrics to understand how closely agents match equilibrium game play and the relative stability or variation in their play over time.  We find in particular that (1) {\em summarized} representations, (2) representations that use {\em regret} rather than raw payoff, and (3) representations that include information only about one's {\em own action} rather than everyone's actions generally lead to the most stable, equilibrium-like behavior. 

A heuristic inspection of the LLM agents' chat histories reveals some intriguing qualitative insights behind these findings.  We briefly discuss these qualitative observations along each axis in our representation framework below. (For more detailed insights about LLM responses under various representations in Games A and B, see Table~\ref{tab:drivers} in the appendix.)
\begin{itemize}
\item {\em Action informativeness.}  Informally, we observed that providing information about everyone's actions could elicit counterfactual, anticipatory reasoning about actions that others could take in future rounds given their historical actions as well as observations about equilibrium and best response; however, the agent might then either proceed to make incorrect inferences with that information, or might fail to utilize that knowledge in subsequent decision-making, resulting in non-equilibrium play. For example, rather than avoiding routes that the agent believes others would switch to in the next round based on historical patterns, the agent might (erroneously) proceed to follow crowd behavior in the next round.  On the other hand, providing the agent with only their own actions seemed to mitigate these types of errors.
\item {\em Reward informativeness.}  In both Games A and B, it appeared that providing agents with regrets caused agents to anchor their decision-making to patterns of low regrets obtained on certain routes.  This type of information about the relative value of the best action helped stabilize game play and keep agents closer to equilibrium.  On the other hand, when provided with raw payoff information, LLM agents seemed to reason about the risk of high congestion (or equivalently, low payoffs); this evaluation of risk led to more ambiguous inferences, increasing both instability and divergence from equilibrium play.
\item {\em Prompting style.}  Finally, while the full-chat prompting style tends to elicit anticipatory behavior in agents, it also could result in incorrect inferences made by agents, often because the agent myopically only accounts for the history information in the immediate preceding round.  This gap meant that even agents' behavior might diverge from equilibrium despite the richer state information. By contrast, summarization preserved accurate inferences from history while encouraging decisions that are closer to the rational best response.
\end{itemize}

Of course, these are only anecdotal observations from our experiments.  Indeed, one concrete direction for future work is to carry out far more sophisticated forensics on the internal behavior of these agents, uncovering the extent to which their manifested behaviors are consistent with behavioral biases observed in real-world human strategic behavior (e.g., \cite{kahneman2013prospect,tversky1974judgment}).  

Our work suggests several other natural and fruitful directions for future exploration.  First, our work focuses on selfish routing games with a specific topology.  We expect that investigating a number of other game classes, and even some of those discussed in prior work (cf.~Section \ref{sec:related}) through the lens of our framework would provide broader evidence for the implications of state representations along the axes we have defined.  Second, our work focuses on a particular model (GPT-4o), and we observe a particular range of quantitative and qualitative model behaviors as a result.  Given the rapid pace of innovation in LLM models and architectures, we think it is natural to expand the line of inquiry we pursue in this paper to newer models, e.g., sequential reasoning models such as OpenAI's o1 and o3, and DeepSeek's R1 (that said, a current challenge here is the significant increase in simulation cost). Finally, as briefly discussed in Section \ref{sec:Framework}, there may be other interesting dimensions of state representations to explore; notably, in more complex games with longer time horizons, it seems useful and interesting to explore the role of curtailing the depth of history provided to the agent.  
\newpage

\bibliographystyle{ACM-Reference-Format}
\bibliography{references}

\newpage

\appendix

\section{LLM Agent Prompts}
\label{sec:prompts}

Figure~\ref{fig:system} below presents the game instructions that we provide to the LLM agents, adapted from the script given to human subjects in \cite{rapoport_choice_2009}.


\begin{figure}[H]
    \centering
    \begin{tcolorbox}[colframe=red!60, colback=red!10, arc=10pt, left=2pt, right=2pt, boxrule=1pt, width=\dimexpr\linewidth-2pt\relax, title={Game Environment (Round $i$)}]
\footnotesize
You will be participating in an experiment on route selection in traffic networks.
During this experiment you'll be asked to make many decisions about route selection in a traffic network game.
Your payoff will depend on the decisions you make as well as the decisions made by the other participants. There are 18 participants in this experiment, including yourself, who will be asked to serve as drivers and choose a route to travel in a traffic network game that is described below.
You will play the game for 40 identical rounds.

Consider the very simple traffic network described below.

\medskip

Nodes:

O L R D

\medskip

Segments and associated costs:

Segment O-L, cost function: 10 * X

Segment O-R, cost function: 210

Segment L-D, cost function: 210

Segment R-D, cost function: 10 * X

\medskip

Each driver is required to choose one of 2 routes to travel from the starting point, denoted by O, to the final destination, denoted by D. There are 2 alternative routes and they are denoted by ['O-L-D', 'O-R-D'].

Travel is always costly in terms of the time needed to complete a segment of the road, tolls, fuel etc. The travel costs are written near each segment of the route you choose. For example, if you choose route O-L-D, you will be charged a total cost of 10X + 210 where X indicates the number of participants who choose segment O-L to travel from O to L plus a fixed cost of 210 for traveling on segment L-D.

Similarly, if you choose route O-R-D, you will be charged a total travel cost of 210 + 10Y, where Y indicates the number of participants who choose the segment R-D to drive from O to D.
Please note that the cost charged for segments O-L and R-D depends on the number of drivers choosing them.

In contrast, the cost charged for traveling on segments L-D and O-R is fixed at 210 and does not depend on the number of drivers choosing them.

All the drivers make their route choices independently of one another and leave point O at the same time.

\medskip

Example.

\smallskip

If you happen to be the only driver who chooses route O-L-D, and all other 17 drivers choose route O-R-D, then your travel cost from point O to point D is equal to (10 x 1) + 210 = 220.
If, on another round, you and 2 more drivers choose route O-R-D and 15 other drivers choose route O-L-D, then your travel cost for that round will be 210 + (10 x 3) = 240.

\medskip

At the beginning of each round, you will receive an endowment of 400 points.
Your payoff for each round will be determined by subtracting your travel cost from your endowment.
Your goal is to maximize your payoff (likewise minimize your cost).
At the end of each round, you will be informed of the number of drivers who chose each route and your payoff for that round. 
All 40 rounds have exactly the same structure.

\medskip

The output should be formatted as a JSON instance that conforms to the JSON schema below.

\medskip

As an example, for the schema \{"properties": \{"foo": \{"title": "Foo", "description": "a list of strings", "type": "array", "items": \{"type": "string"\}\}\}, "required": ["foo"]\}
the object \{"foo": ["bar", "baz"]\} is a well-formatted instance of the schema. The object \{"properties": \{"foo": ["bar", "baz"]\}\} is not well-formatted.

\medskip

Here is the output schema:

\medskip

```
\{"properties": \{"route": \{"title": "Route", "description": "choice of route", "type": "string"\}\}, "required": ["route"]\}
```
    \end{tcolorbox}
    \caption{Game description (adapted from \cite{rapoport_choice_2009}) and formatting instructions given with every API call. Payoff example given is adjusted according to whether agents are playing Game A or Game B. Formatting instructions are generated by LangChain.}
    \label{fig:system}
    \Description{}
\end{figure}

\newpage
\section{Baseline Algorithms and Performance}
\label{sec:rl_algs}

\subsection{EXP3}
The EXP3 (Exponential-weight algorithm for Exploration and Exploitation) algorithm, introduced in \cite{auer_nonstochastic_2002}, is designed for multi-armed bandit settings where losses or rewards are determined by an arbitrary (possibly adversarial) environment. The algorithm proceeds as follows.
Given a set of $k$ arms, let $w_i(t)$ denote the weight assigned to arm $i$ at time $t$. The steps are:
\begin{itemize}
    \item Set initial weights $w_i(1)=1$ for all arms $i\in[k]$
    \item Compute the probability of selecting each arm $i$ as
    \begin{equation}
        p_i(t)=(1-\gamma)\frac{w_i(t)}{\sum_{j=1}^k}w_j(t)+\frac{\gamma}{k}
    \end{equation}
    where $\gamma\in[0,1]$ is the exporation parameter; we use $\gamma=0.75$ for the purposes of this paper.
    \item Sample an arm $i_t$ according to the probability distribution $p(t)$
    \item Observe the loss $\ell_{i_t}(t)$ and define an unbiased loss estimate as
    \begin{equation}
    \hat{\ell}_{i_t}(t)=\frac{\ell_{i_t}(t)}{p_{i_t}(t)}
    \end{equation}
    \item Update the weights according to
    \begin{equation}
        w_i(t+1)=w_i(t)\exp(-\eta\hat{\ell}_i(t))
    \end{equation}
    where $\eta > 0$ is the learning rate. We set $\eta=0.75$, the same as $\gamma$.
    \item Repeat for $T$ rounds.
\end{itemize}

\subsection{Multiplicative Weights Update}
The Multiplicative Weights Update (MWU) algorithm was introduced in \cite{freund_adaptive_1999}. It is a widely used technique for learning in repeated decision-making scenarios. MWU differs from EXP3 in that it has access to the losses of every action, opposed to only the chosen action. The algorithm is as follows.
Let $k$ denote the number of actions, and let each action $i\in[k]$ have corresponding weight $w_i(t)$. The steps are:
\begin{itemize}
    \item Set initial weights $w_i(1)=1$ for all $i\in[k]$.
    \item Compute the probability of selecting each action as
    \begin{equation}
    p_i(t)=\frac{w_i(t)}{\sum_{j=1}^kw_j(t)}
    \end{equation}
    \item Choose an action $i_t$ according to the probability distribution $p(t)$.
    \item Observe the loss $\ell_i(t)$ for each action $i$
    \item Update the weights according to
    \begin{equation}
    w_i(t+1)=w_i(t)\cdot e^{-\eta\ell_i(t)}
    \end{equation}
    where $\eta>0$ is the learning rate. We set $\eta=0.75$.
    \item Repeat for $T$ rounds.
\end{itemize}

\newpage
\subsection{Temporal Performance of EXP3 and MWU}

\begin{figure}[ht]
    \centering
    \begin{subfigure}{0.48\textwidth}
        \centering
        \includegraphics[width=\linewidth]{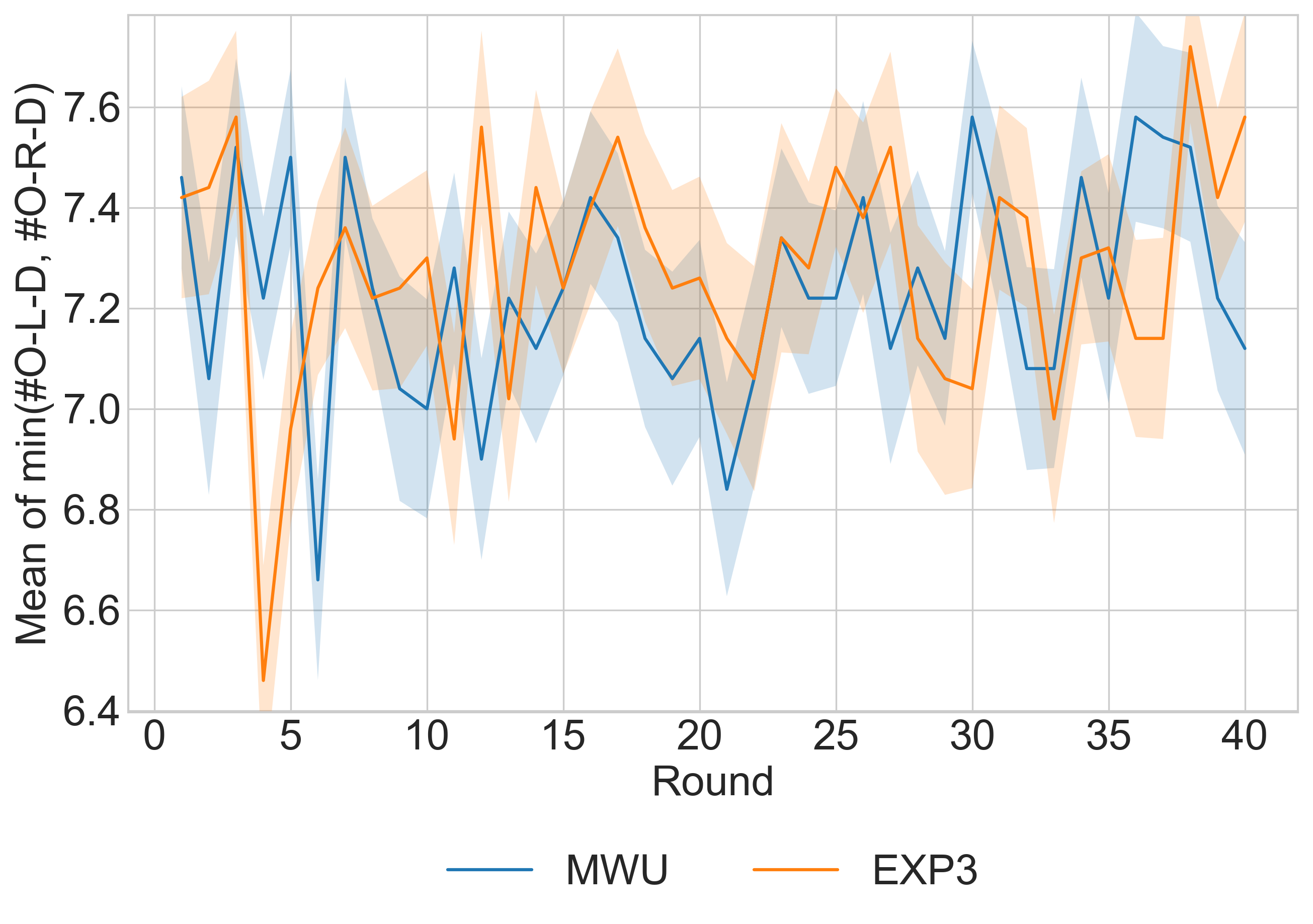}
        \caption{Fig.~\ref{fig:RL-route-trends-A}. Mean route choice per agent, Game A}
        \label{fig:RL-route-trends-A}
    \end{subfigure}
    \hfill
    \begin{subfigure}{0.48\textwidth}
        \centering
        \includegraphics[width=\linewidth]{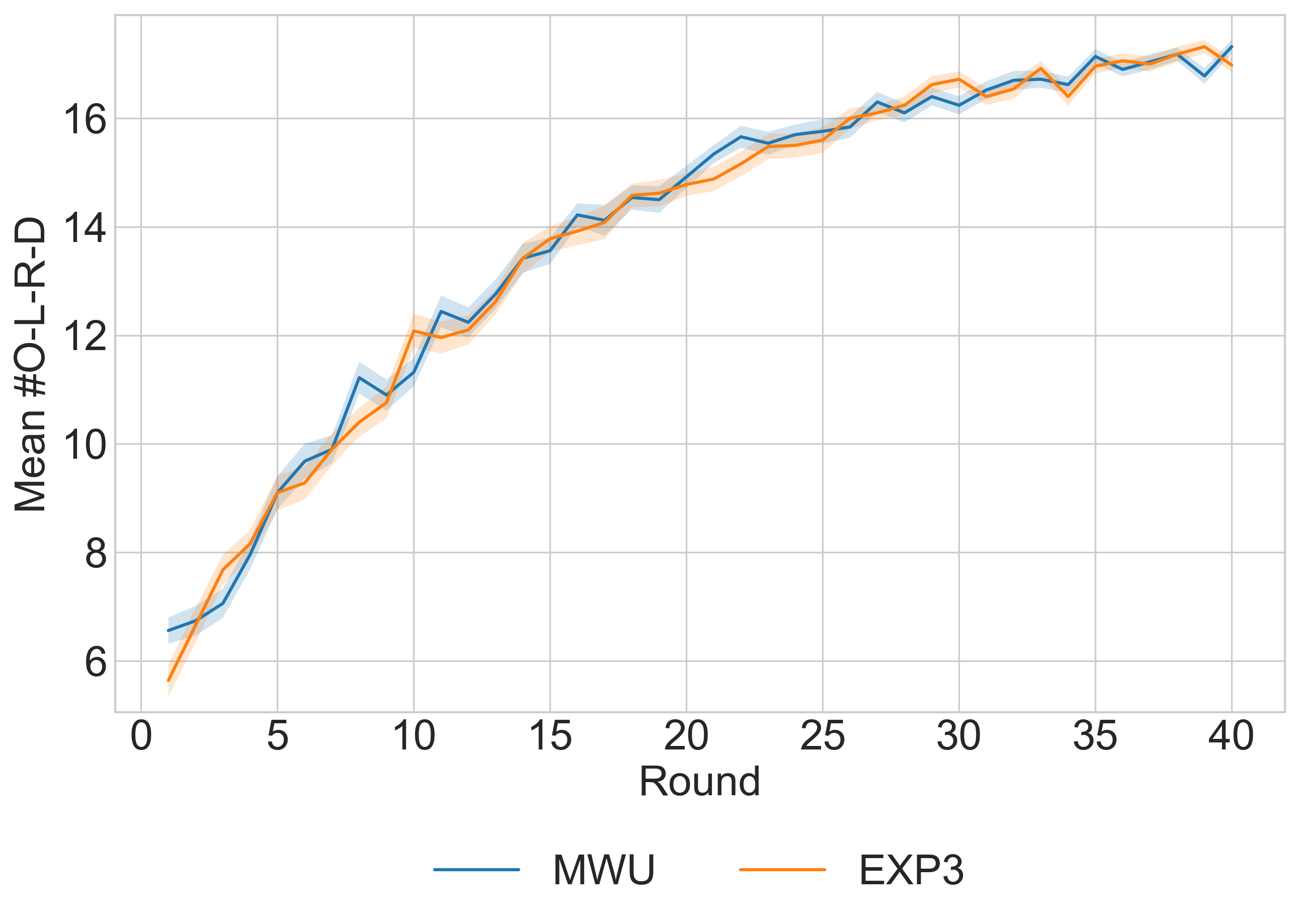}
        \caption{Fig.~\ref{fig:RL-route-trends-B}. Mean route choice per agent, Game B}
        \label{fig:RL-route-trends-B}
    \end{subfigure}
    \label{fig:RL-route-trends}
    \Description{}
\end{figure}

\begin{figure}[ht]
    \centering
    \begin{subfigure}{0.48\textwidth}
        \centering
        \includegraphics[width=\linewidth]{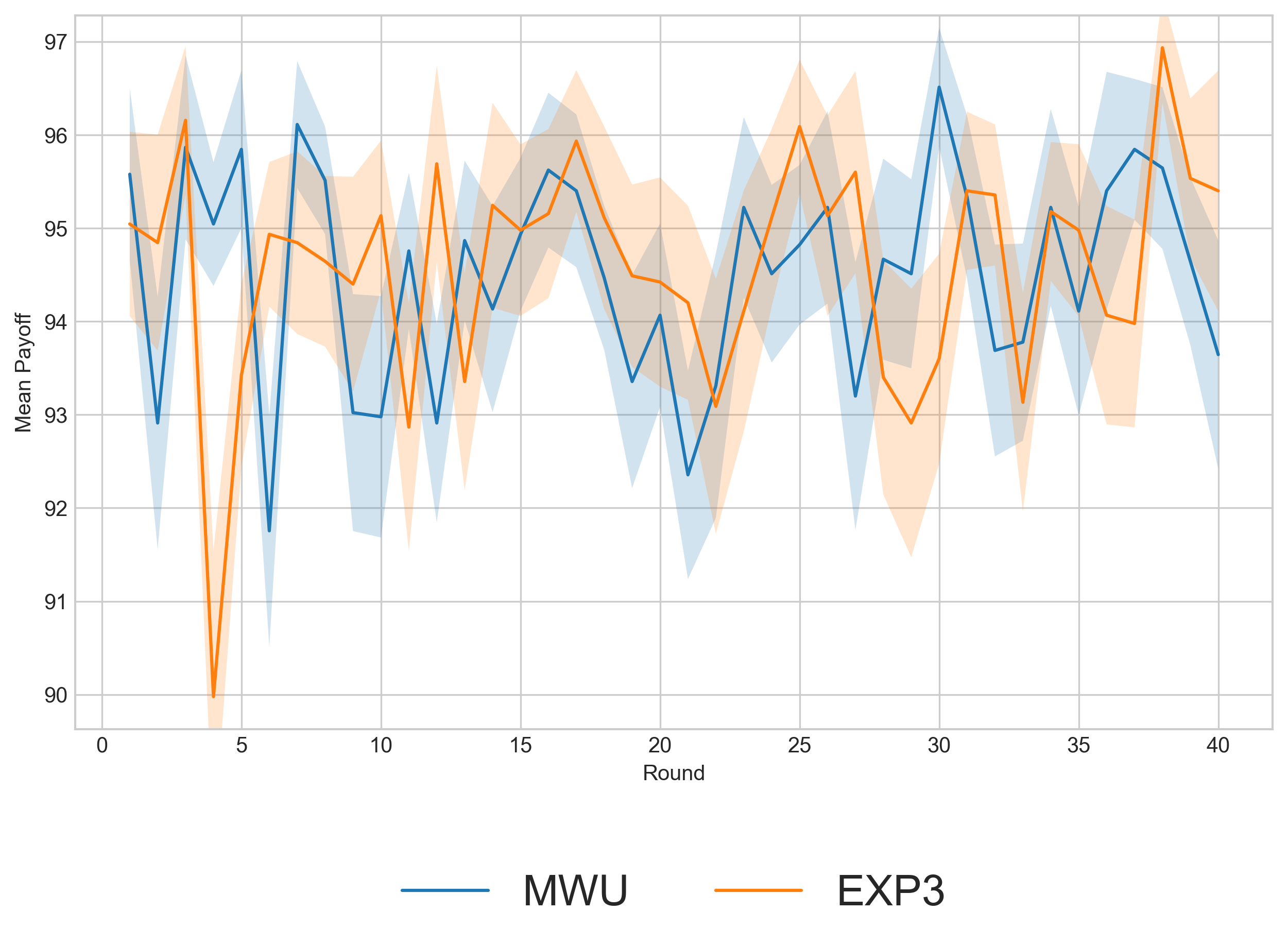}
        \caption{Fig.~\ref{fig:RL-reward-trends-A}. Mean payoff per agent, Game A}
        \label{fig:RL-reward-trends-A}
    \end{subfigure}
    \hfill
    \begin{subfigure}{0.48\textwidth}
        \centering
        \includegraphics[width=\linewidth]{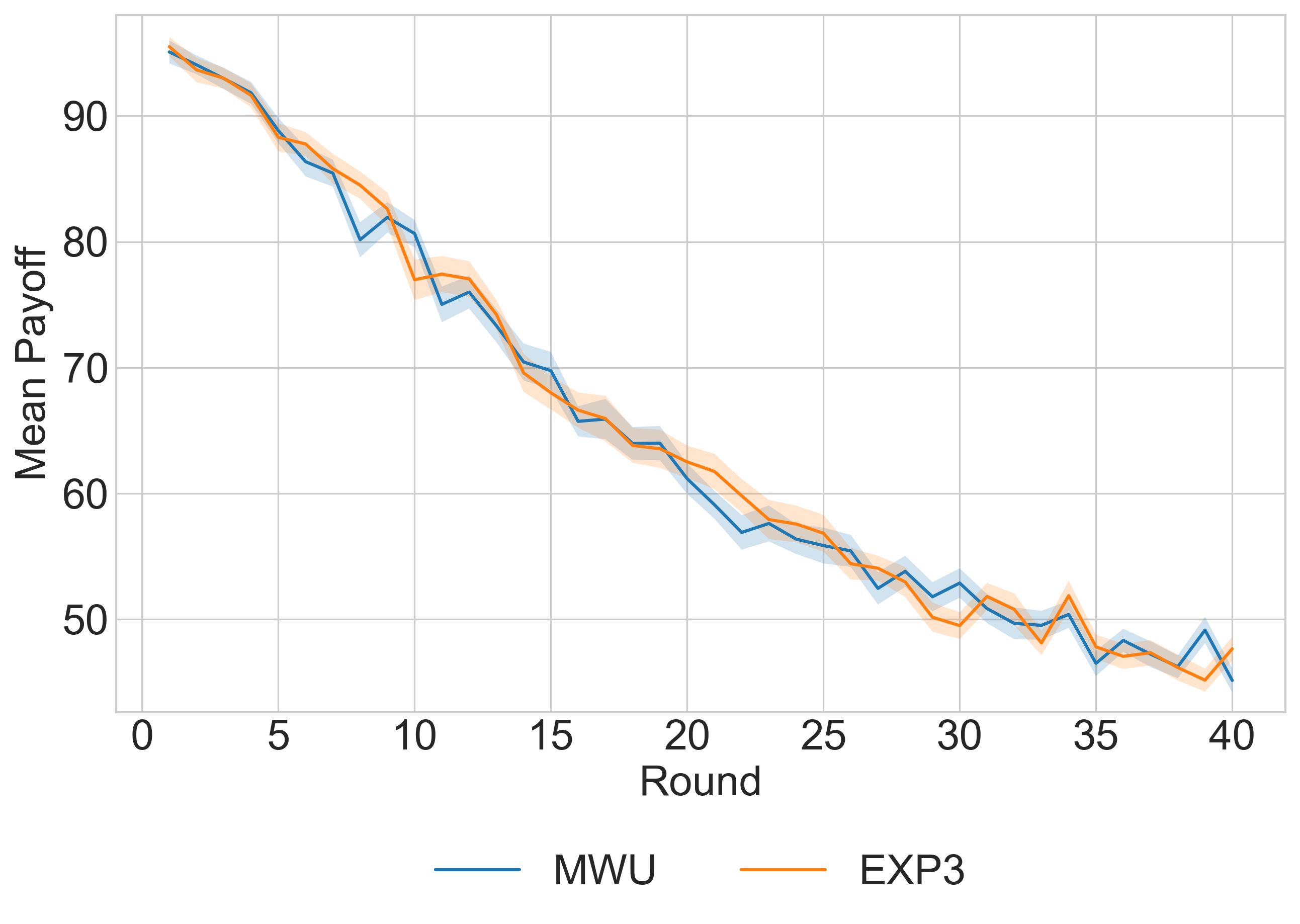}
        \caption{Fig.~\ref{fig:RL-reward-trends-B}. Mean payoff per agent, Game B}
        \label{fig:RL-reward-trends-B}
    \end{subfigure}
    \label{fig:RL-reward-trends}
    \Description{}
\end{figure}

\begin{figure}[ht]
    \centering
    \begin{subfigure}{0.48\textwidth}
        \centering
        \includegraphics[width=\linewidth]{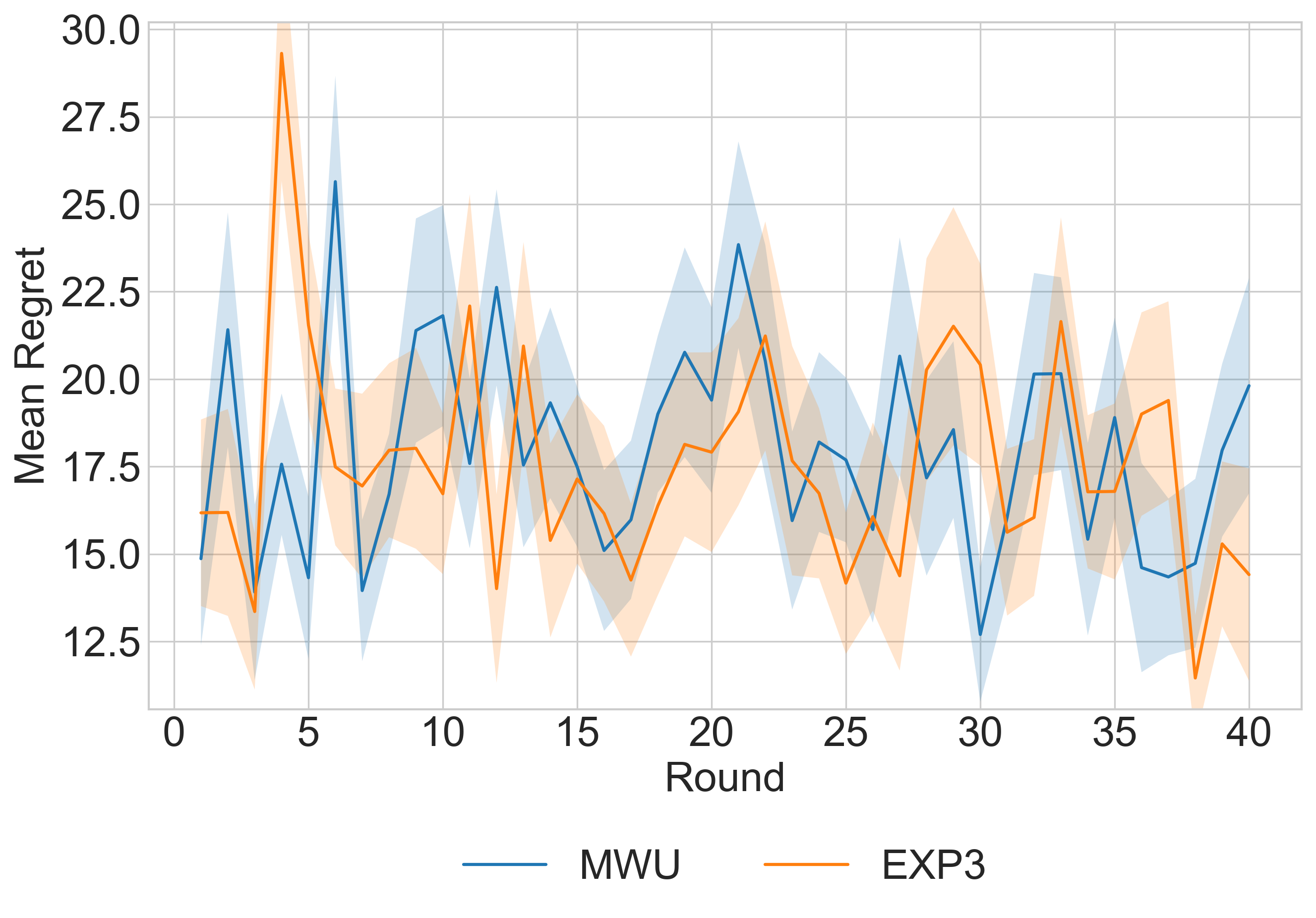}
        \caption{Fig.~\ref{fig:RL-regret-trends-A}. Mean regret per agent, Game A}
        \label{fig:RL-regret-trends-A}
    \end{subfigure}
    \hfill
    \begin{subfigure}{0.48\textwidth}
        \centering
        \includegraphics[width=\linewidth]{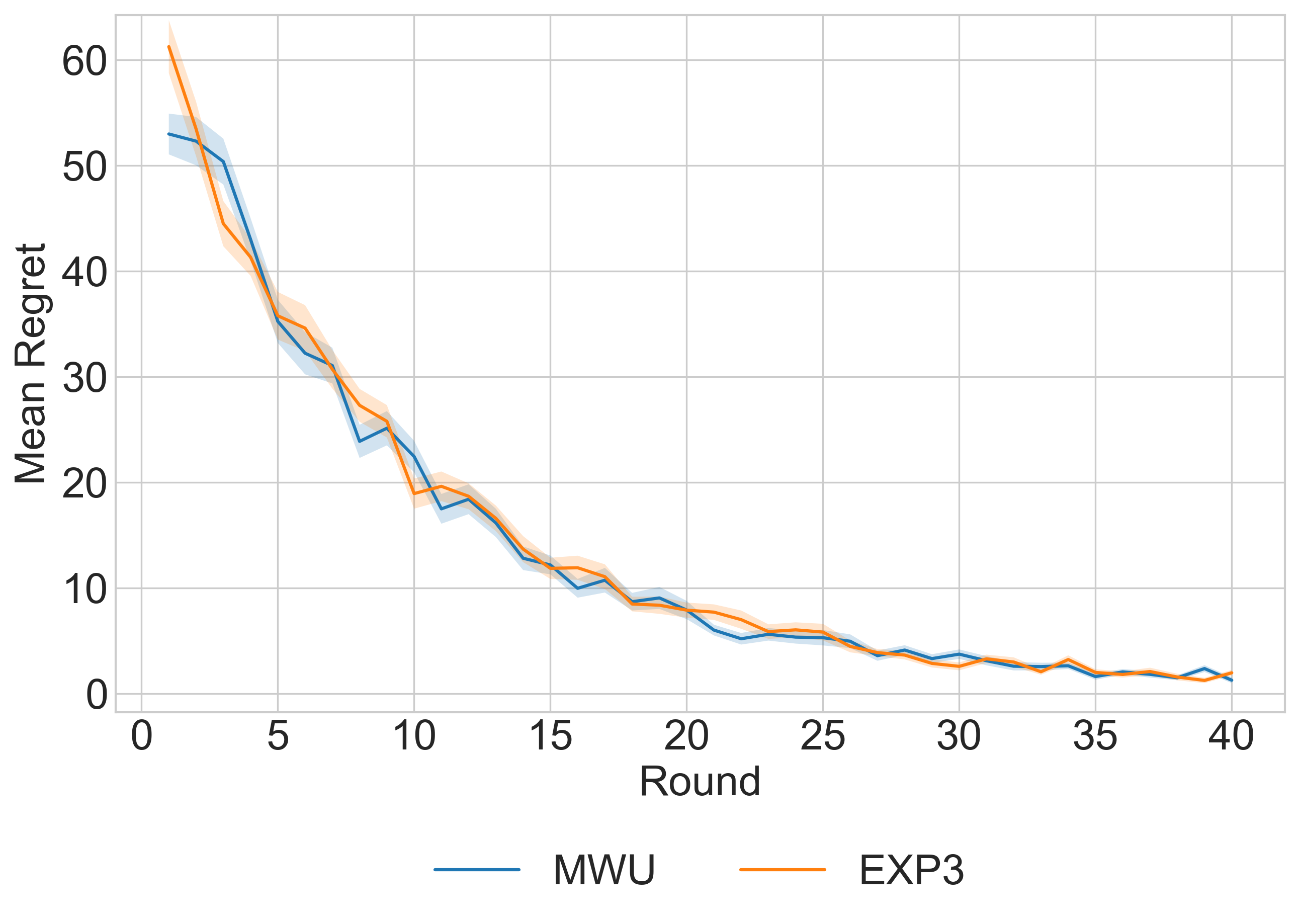}
        \caption{Fig.~\ref{fig:RL-regret-trends-A}. Mean regret per agent, Game B}
        \label{fig:RL-regret-trends-B}
    \end{subfigure}
    \label{fig:RL-regret-trends}
    \Description{}
\end{figure}

\begin{figure}[ht]
    \centering
    \begin{subfigure}{0.48\textwidth}
        \centering
        \includegraphics[width=\linewidth]{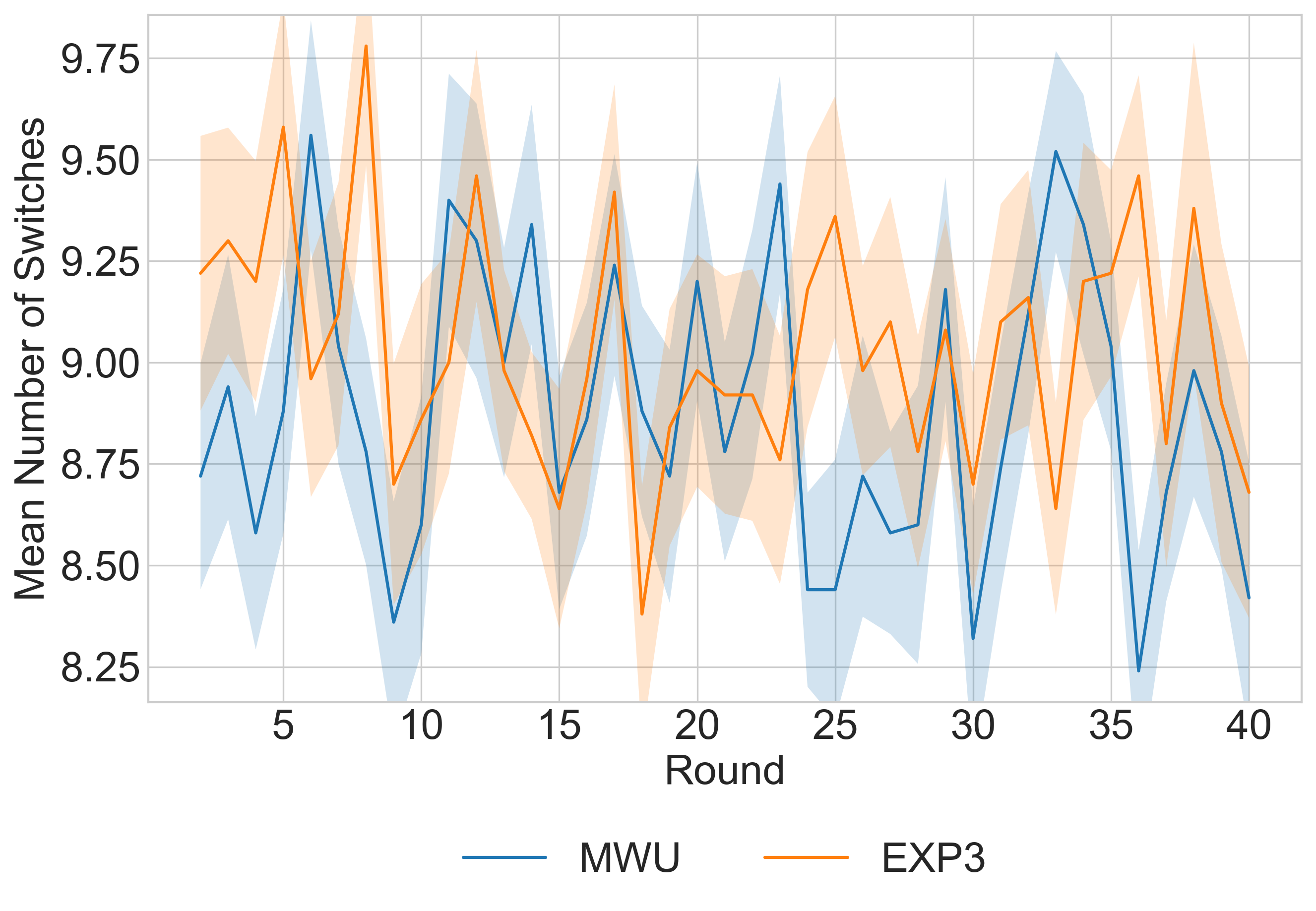}
        \caption{Fig.~\ref{fig:RL-switch-trends-A}. Mean number of switches per agent, Game A}
        \label{fig:RL-switch-trends-A}
    \end{subfigure}
    \hfill
    \begin{subfigure}{0.48\textwidth}
        \centering
        \includegraphics[width=\linewidth]{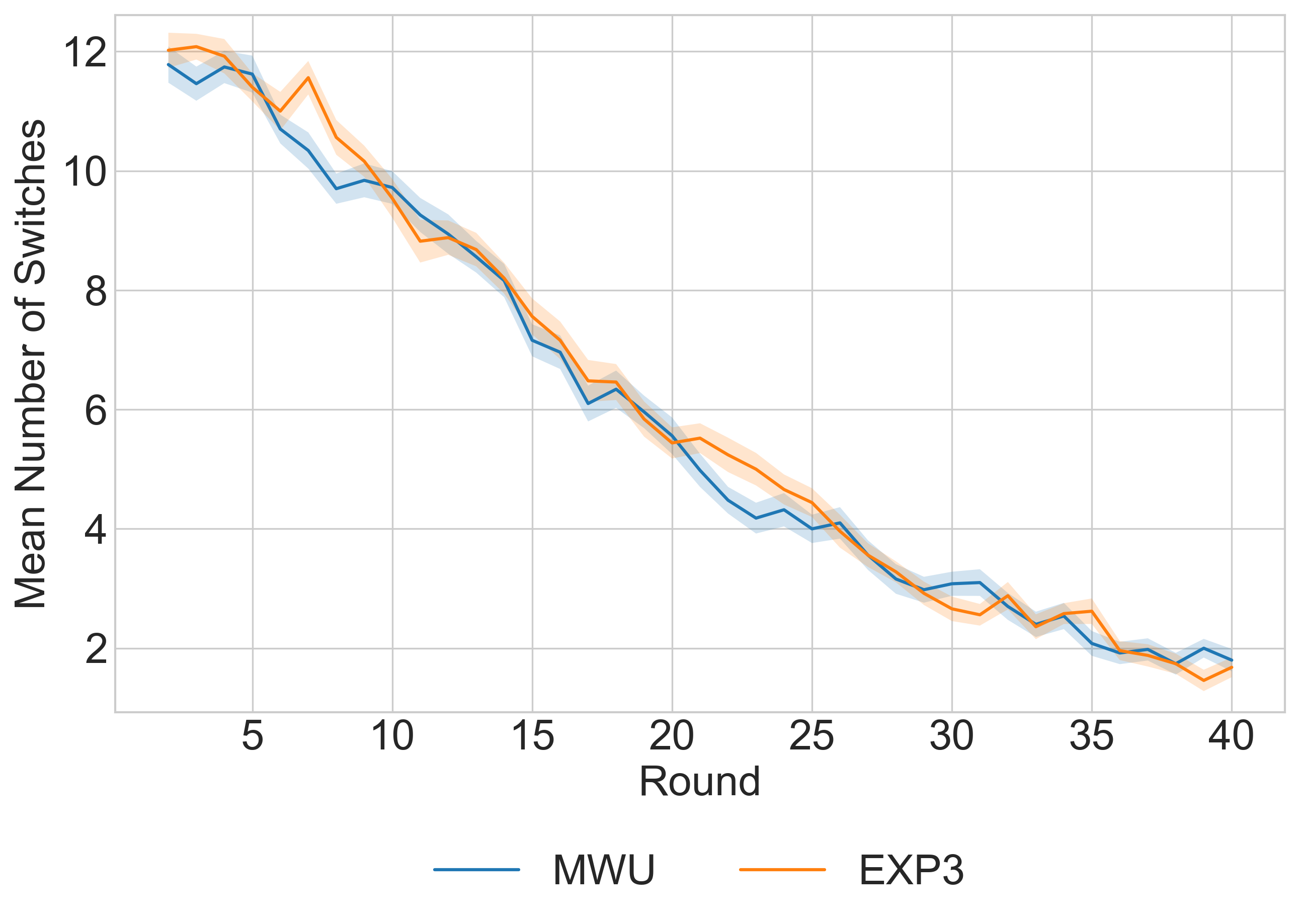}
        \caption{Fig.~\ref{fig:RL-switch-trends-B}. Mean number of switches per agent, Game B}
        \label{fig:RL-switch-trends-B}
    \end{subfigure}
    \label{fig:RL-switch-trends}
    \Description{}
\end{figure}

\newpage

\section{Qualitative Analysis of LLM Output}
\label{sec:qualitative-analysis-table}

\begin{table}[ht]
    \footnotesize
    \caption{Main drivers of decisions from LLM output in Games A and B for each state representation. These drivers are obtained from analyzing the LLM outputs for one agent in the last round of one trial of each game under each representation. The LLMs are prompted to provide their reasoning for their decisions through explicit instructions in the prompt to think step-by-step in a chain-of-thought manner.}
    \label{tab:drivers}
    \begin{tabular}{  l  p{6cm}  p{6cm} }
        \toprule
        & \textbf{Game A}   
        & \textbf{Game B} \\\midrule
        F-PO &
        Given the high congestion last round on O-L-D, \textbf{anticipates} many drivers may choose to switch to O-R-D in search of a less crowded route, but \textbf{mistakenly} decides to switch to O-R-D, predicting others will move away from O-L-D. &
        \textbf{Anticipates route dynamics} based on \textbf{potential switching by others after experiencing congestion}. Considers forecasting the movements of others in strategic considerations, but does not explain how.
        \\\hline
        S-PO &
        Choosing O-R-D is likely to result in a \textbf{consistent payoff} based on previous rounds. This route offers less risk of extremely low payoffs compared to O-L-D, since your payoff has been \textbf{more predictive} in these later rounds. &
        \textbf{Route analysis without regard for other's actions}. Decision strategy based on maximizing payoffs based on historical \textbf{patterns} of payoffs within routes.
        \\\hline
        F-RO &
        Recognizes that drivers respond to high regrets by \textbf{switching} routes, and thus fluctuating popularity between routes. \textbf{Anticipates} fewer drivers on O-R-D following its high regret cost in the previous round. &
        Considers the \textbf{consistency} of \textbf{zero regret} on O-L-R-D. Notes that it will keep observing for any changes in route dynamics, as these could affect future decisions.
        \\\hline
        S-RO &
        Sticks with O-R-D since it has \textbf{consistently} resulted in \textbf{minimal regret}. The stability in regret suggests that the \textbf{dynamics} among participants have settled into a \textbf{pattern} where O-R-D is preferable for minimizing costs. &
        Finds \textbf{pattern of no regret} on O-L-R-D, and suboptimal regret with other route choices historically.
        \\\hline
        F-PE &
        Observes that \textbf{congestion tends to shift} between routes, and thus infers choosing the \textbf{currently unused route} is likely to be beneficial. Switching to O-L-D should allow you to capitalize on \textbf{minimal congestion} and result in better payoff through lower costs. &
        \textbf{Avoids routes that start with O-L due to significant congestion}. Infers that because no drivers were on O-R-D, it potentially has lower congestion.
        \\\hline
        S-PE &
        Trades off between noting there is an \textbf{equilibrium} around 9 participants (where payoffs between routes are balanced), and that O-R-D experienced \textbf{high congestion} in recent rounds, making it less desirable. &
        Considers \textbf{congestion risk if everyone thinks similarly} and chooses O-L-R-D; considers \textbf{switching to less populated routes} if more participants begin moving to O-L-R-D. Chooses route that \textbf{balances acceptable payoff with congestion risk}.
        \\\hline
        F-RE &
        Given the \textbf{higher costs} on O-R-D, it reasons that it's reasonable to \textbf{expect that some drivers will switch} to O-L-D in order to minimize costs. However, it \textbf{mistakenly} believes that it makes sense for you to \textbf{switch} to O-L-D this round, aligning with the expected trend. &
        Evaluates route dynamics and \textbf{anticipates} driver behavior. Sticks with O-L-R-D due to its proven \textbf{consistency} in delivering \textbf{zero-regret} outcomes and cost-efficient performance.  
        \\\hline
        S-RE &
        The strategic choice would be to select the route that has \textbf{recently shown minimal regret} under varying conditions, primarily when drivers are \textbf{distributed more evenly}. Choice relies on past observations where O-R-D \textbf{consistently} resulted in minimal regret. &
        \textbf{Every instance} of choosing O-L-R-D has resulted in \textbf{no regret}. Given the \textbf{complete convergence} to O-L-R-D \textbf{among participants} in \textbf{recent rounds}, \textbf{switching routes introduces unnecessary risk}.\\
    \bottomrule
    \end{tabular}
\end{table}

\end{document}